\documentclass[sigconf]{acmart}

\AtBeginDocument{%
  }

\usepackage{amsmath}
\allowdisplaybreaks

\usepackage{amssymb}
\usepackage{mathtools}
\usepackage{bm}  
\usepackage{soul}
\usepackage{enumitem}
\usepackage{booktabs}
\usepackage{multirow}
\usepackage{makecell}
\usepackage{graphicx}  
\usepackage{caption}   
\usepackage[table]{xcolor}
\usepackage{float}
\usepackage{needspace}

\copyrightyear{2026}
\acmYear{2026}
\setcopyright{cc}
\setcctype{by}
\acmConference[KDD '26]{Proceedings of the 32nd ACM SIGKDD Conference on Knowledge Discovery and Data Mining V.2}{August 09--13, 2026}{Jeju Island, Republic of Korea}
\acmBooktitle{Proceedings of the 32nd ACM SIGKDD Conference on Knowledge Discovery and Data Mining V.2 (KDD '26), August 09--13, 2026, Jeju Island, Republic of Korea}
\acmDOI{10.1145/3770855.3818141}
\acmISBN{979-8-4007-2259-2/2026/08}

\begin{document}





\title{Geometry-First Generative Spatial Single-Cell Reconstruction}


\author{Ehtesamul Azim}
\affiliation{%
  \institution{University of Central Florida}
  \city{Orlando}
  \state{Florida}
  \country{USA}}
\email{ehtesamul.azim@ucf.edu}

\author{Muhtasim Noor Alif}
\affiliation{%
  \institution{University of Central Florida}
  \city{Orlando}
  \state{Florida}
  \country{USA}}
\email{muhtasimnoor.alif@ucf.edu}

\author{Tae Hyun Hwang}
\affiliation{%
  \institution{Vanderbilt University Medical Center}
  \city{Nashville}
  \state{Tennessee}
  \country{USA}}
\email{taehyun.hwang@vumc.org}

\author{Yanjie Fu}
\affiliation{%
  \institution{Arizona State University}
  \city{Tempe}
  \state{Arizona}
  \country{USA}}
\email{yanjie.fu@asu.edu}

\author{Wei Zhang}
\authornote{Corresponding author.}
\affiliation{%
  \institution{University of Central Florida}
  \city{Orlando}
  \state{Florida}
  \country{USA}}
\email{wzhang.cs@ucf.edu}

\renewcommand{\shortauthors}{Ehtesamul Azim, Muhtasim Noor Alif, Tae Hyun Hwang, Yanjie Fu, \& Wei Zhang}

\begin{abstract}
Single-cell RNA sequencing (scRNA-seq) profiles large numbers of cells but loses spatial context, whereas spatial transcriptomics (ST) preserves partial spatial structure at lower resolution. Most existing integration methods either deconvolve spot mixtures or map cells onto a measured spot lattice, which ties reconstructions to a fixed grid and slide-specific coordinate systems, a limitation that is especially problematic in unpaired settings. We propose \textbf{GEARS}, a geometry-first framework that reconstructs an \emph{intrinsic} single-cell spatial geometry guided by ST, without relying on cell-type labels, histological images, or cell-to-spot assignment. GEARS first learns a domain-invariant expression encoder that aligns ST spots and dissociated cells, and then trains a permutation-equivariant generator with a diffusion-based refiner with EDM-style preconditioning to generate local spatial geometries under pose-invariant supervision derived from ST coordinates. At inference, GEARS reconstructs geometry on many overlapping subsets of scRNA-seq cells, aggregates predicted pairwise distances across subsets, and solves a global distance-geometry problem to obtain canonical two-dimensional coordinates and a dense distance matrix. Extensive quantitative and qualitative experiments, including cross-section generalization, show that GEARS consistently improves global distance preservation, local neighborhood fidelity, and spatial distribution alignment compared to strong spatial mapping and deconvolution baselines. 
\let\thefootnote\relax\footnotetext{Release code and preprocessed data can be found at \href{https://github.com/ehtesam3154/GEARS}{github.com/ehtesam3154/GEARS}}
\end{abstract}






\maketitle
\vspace{-0.2cm}
\section{Introduction}
Modern single-cell RNA sequencing (scRNA-seq) allows us to measure gene activity for hundreds of thousands of individual cells in a tissue~\cite{picelli2014full, zheng2017massively}. However, this powerful technology comes with a major limitation: during the sequencing process, cells are physically separated from their original tissue, and all spatial information is lost~\cite{longo2021integrating}. As a result, scRNA-seq data tell us what each cell is doing, but not where the cell was located. Spatial location is critical for understanding how cells interact, how tissues are organized, and how diseases such as cancer develop. Recovering the spatial arrangement of single cells from scRNA-seq data has therefore become an important and challenging computational problem~\cite{biancalani2021deep, cang2020inferring, azim2025biological}.

Spatial transcriptomics (ST) technologies partially address this limitation by measuring gene expression together with spatial coordinates, thereby preserving the physical structure of a tissue~\cite{marx2021method, crosetto2015spatially, asp2020spatially, rao2021exploring}. However, current ST technologies typically operate at lower resolution than scRNA-seq: each spatial location captures signals from multiple cells, and the number of spatial spots is limited. These complementary strengths and weaknesses motivate the problem of single-cell spatial reconstruction, which aims to infer plausible 2D locations for scRNA-seq cells by leveraging spatial information from an ST reference~\cite{biancalani2021deep, cang2020inferring, moriel2021novosparc}. In practice, many scRNA-seq datasets lack matched spatial measurements, and researchers often rely on an ST atlas or dataset from the same tissue type as reference rather than paired measurement from the same tissue section or individual. \textit{Because large collections of scRNA-seq datasets have already been generated across tissues, conditions, and disease states, effective 2D reconstruction makes it possible to spatially reinterpret these existing data without requiring new experiments}~\cite{schaum2018single, the2022tabula}. This capability substantially extends the value of current single-cell(SC) datasets by enabling spatial analyses, such as tissue organization and local cell interactions, which were previously inaccessible.

Existing approaches to single-cell spatial reconstruction face substantial challenges in this unpaired setting, where scRNA-seq and ST data are collected from different tissue samples or individuals~\cite{stuart2019comprehensive}. Differences in experimental protocols, gene coverage, and biological states introduce significant domain shift between datasets, making direct matching unreliable. Most methods rely on a set of shared genes to relate the two modalities, which amplifies the impact of noise and batch effects when no paired measurements are available for calibration~\cite{haghverdi2018batch}. In addition, variation in cell-type composition and abundance across samples can lead to ambiguous mappings, as multiple spatial configurations may be equally consistent with the reference data.

More fundamentally, existing methods implicitly assume that single cells should be placed within the 2D coordinate system of the reference spatial dataset. This assumption is often violated in unpaired settings. Even for the same tissue type, different samples can exhibit substantially different global organization, local structures, and spatial patterns~\cite{khan2025comprehensive}. Forcing single cells to align with absolute coordinates of a reference tissue can therefore introduce systematic bias~\cite{zhao2021spatial} and distort true spatial relationships among cells. \textit{As a result, existing methods conflate learning transferable spatial principles with reproducing a specific tissue geometry.} These limitations motivate new models that treat single-cell spatial organization as a latent structure, guided, but not constrained, by reference spatial data, and that are robust and scalable in unpaired settings.

\textbf{Our contribution: Geometry-aware reconstruction via pose-invariant supervision and distance-first inference.}
\ul{We treat spatial reconstruction as \emph{geometry generation}: learning a continuous spatial organization of cells from expression, using ST only as geometric supervision.}
Unlike existing approaches that tie reconstructions to a measured lattice and impose assignment/composition assumptions, our method produces a sample-specific spatial layout whose correctness is defined by geometric consistency (global distances, local neighborhoods, and multiscale structure) rather than by matching to a fixed grid. We learn a shared expression representation that aligns spatial and dissociated profiles, then train a permutation-equivariant set model to generate geometry under pose-invariant targets derived from spatial coordinates. Finally, we scale reconstruction to large single-cell cohorts by predicting geometry on many overlapping subsets and assembling a coherent global reconstruction by stitching \emph{distance constraints}. Our key contributions are as follows:
\begin{itemize}[leftmargin=*, itemsep=1pt]
    \item[$\bullet$] \textit{Geometry-first spatial reconstruction:} we formulate reconstruction as generating a continuous spatial geometry for single cells from expression, avoiding cell-to-spot assignment and spot-level deconvolution assumptions.
    \item[$\bullet$] \textit{Pose-invariant supervision from ST:} we learn from spatial coordinates through intrinsic geometric structure rather than absolute positions, reducing dependence on slide-specific coordinate frames and supporting cross-section generalization.
    \item[$\bullet$] \textit{Permutation-equivariant generator--refiner model:} we use a set-structured architecture with a coarse geometric proposal and diffusion-based refinement that improves global consistency and calibrates spatial scale.
    \item[$\bullet$] \textit{Patchwise distance-first inference:} we scale to large scRNA-seq cohorts by reconstructing overlapping subsets and stitching them by aggregating distance constraints into a global reconstruction.
\end{itemize}

\begin{figure*}[!h]
\vspace{-0.2cm}
    \centering
    \includegraphics[width=1.0\linewidth]{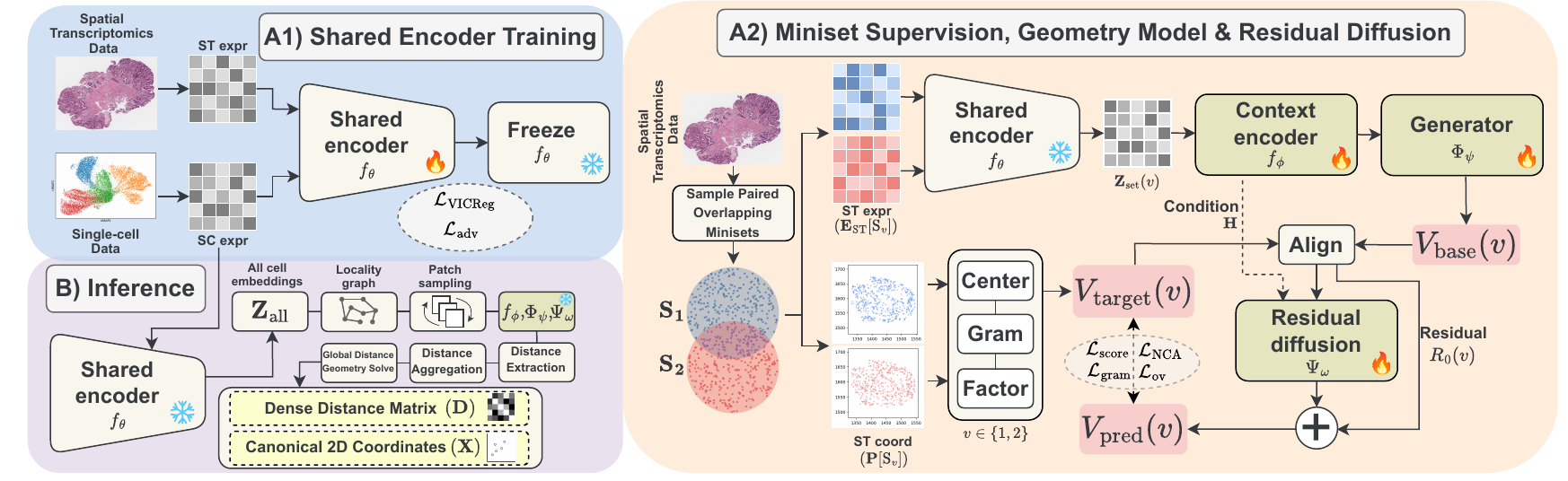}
    
    \vspace{-0.5cm}
    \captionsetup{justification=centering, font=small}
    \caption{
    \textbf{GEARS framework overview.}
    (A1) An encoder is trained to align ST and SC expression into a domain-invariant embedding space.
    (A2) From each ST slide we sample paired, overlapping minisets and train a permutation-equivariant generator with an EDM-preconditioned residual diffusion refiner to match pose-invariant Gram geometry targets.
    (B) At inference, we encode all scRNA-seq cells, sample overlapping patches, predict per-patch geometries, stitch them by aggregating distances, and solve a global distance-geometry problem to output canonical coordinates $\mathbf{X}$ and dense distances $\mathbf{D}$.
    }
    \vspace{-0.5cm}
    \label{fig:framework}
\end{figure*}
\vspace{-0.2cm}
\section{Problem Statement}
\label{prelim}

We address \emph{spatial reconstruction of dissociated single cells} using ST as geometric supervision, without performing cell-to-spot assignment or spot-level deconvolution. Our data consist of $S$ ST slides $\{(\mathbf{E}_{\text{ST}}^{(s)},\mathbf{P}^{(s)})\}_{s=1}^S$, where $\mathbf{E}_{\text{ST}}^{(s)}\in\mathbb{R}^{N_s\times G}$ denotes spot expression over $G$ genes and $\mathbf{P}^{(s)}\in\mathbb{R}^{N_s\times 2}$ denotes measured coordinates, together with an scRNA-seq dataset providing expression only, $\mathbf{E}_{\text{SC}}\in\mathbb{R}^{N_{SC}\times G}$. We write $\mathbf{x}\in\mathbb{R}^{G}$ for a single expression profile (row of $\mathbf{E}_{\text{ST}}$ or $\mathbf{E}_{\text{SC}}$).\footnote{For readability, we omit the slide index $(s)$ when it is not essential and write $(\mathbf{E}_{\text{ST}},\mathbf{P})$ for an ST slide; all definitions apply per slide.} The goal is to reconstruct a coherent \emph{intrinsic} geometry over $N_{SC}$ single cells that is guided by tissue structure observed in ST, such that relative distances and neighborhood relationships are meaningful and transferable across samples. We summarize the reconstruction as a pairwise distance matrix $\mathbf{D}\in\mathbb{R}^{N_{SC}\times N_{SC}}$, where $D_{ij}$ captures reconstructed spatial proximity between cells $i$ and $j$. In practice, we first stitch sparse distances on a global edge set and realize a consistent $\mathbf{X}\in\mathbb{R}^{N_{SC}\times 2}$ from the measurements, from which $\mathbf{D}$ is obtained as all-pairs Euclidean distances.

\vspace{-0.2cm}
\section{Methodology}
\noindent\textbf{Framework overview.}
Fig.~\ref{fig:framework} illustrates \textbf{GEARS} (\underline{\textbf{GE}}ometry-\underline{\textbf{A}}ware \underline{\textbf{R}}econstruction of \underline{\textbf{S}}ingle cells), our proposed model for spatial reconstruction of dissociated scRNA-seq cells using ST as geometric supervision.
We first train a domain-invariant shared expression encoder to align ST spots and dissociated cells in a common embedding space.
Using spatial structure from ST only to define pose-invariant geometric targets, we sample local minisets and train a permutation-equivariant geometry generator with an EDM-preconditioned diffusion refiner~\cite{karras2022elucidatingdesignspacediffusionbased} under Gram-based supervision. At inference, we generate geometries on many overlapping patches of scRNA-seq cells, stitch them by aggregating predicted pairwise distances, and solve a global distance-geometry problem to obtain canonical coordinates $\mathbf{X}$ and a dense distance matrix $\mathbf{D}$.

\vspace{-0.3cm}

\subsection{Domain-Invariant Expression Embeddings}
\textbf{Why learn shared embeddings without spatial coordinates?}
Most spatial--single-cell integration methods either deconvolve spot mixtures or map cells onto a measured spatial lattice, coupling the model to slide-specific coordinate frames that quantize fine structure and hinder transfer across samples. We instead learn a modality-shared expression embedding that aligns spatial and dissociated profiles while suppressing domain-specific technical artifacts, decoupling biology from coordinates. This coordinate-free embedding provides a common representation for downstream geometry learning on ST and expression-conditioned geometric reconstruction of dissociated cells.

\noindent\textbf{\ul{\emph{Modality-invariant biological signal extraction.}}} We train a shared encoder $f_\theta:\mathbb{R}^G\!\to\!\mathbb{R}^h$ that maps gene expression from both ST spots and dissociated cells into a common embedding space, where $h$ is the embedding dimension. Our goal is to preserve biological variation while removing modality and batch-specific effects. We combine VICReg~\cite{bardes2021vicreg} with adversarial domain alignment via gradient reversal (GRL)~\cite{JMLR:v17:15-239}.

\noindent\textit{VICReg on coordinate-free augmentations.}
For each sampled expression profile $\mathbf{x}$, we generate two augmented views $\tilde{\mathbf{x}}^{(1)},\tilde{\mathbf{x}}^{(2)}$ using stochastic gene dropout, additive Gaussian noise, and multiplicative scale jitter. Let $\mathbf{z}_1=f_\theta(\tilde{\mathbf{x}}^{(1)})$ and $\mathbf{z}_2=f_\theta(\tilde{\mathbf{x}}^{(2)})$, stacked over a minibatch of size $B$ as $\mathbf{z}_1,\mathbf{z}_2\in\mathbb{R}^{B\times h}$. We optimize the VICReg objective, where $s_j(\mathbf{z})=\sqrt{\operatorname{Var}(\mathbf{z}_{:,j})}$ is the batch standard deviation of dimension $j$ and $\operatorname{Cov}(\mathbf{z})$ is the batch covariance:
\begin{equation}
\begin{aligned}
\mathcal{L}_{\text{VICReg}} &= \lambda_{\text{inv}} \lVert \mathbf{z}_1 - \mathbf{z}_2 \rVert_F^2 \\
&\quad + \lambda_{\text{var}} \sum_{j=1}^h \Big[\operatorname{ReLU}\!\big(\gamma - s_j(\mathbf{z}_1)\big) + \operatorname{ReLU}\!\big(\gamma - s_j(\mathbf{z}_2)\big)\Big] \\
&\quad + \lambda_{\text{cov}} \sum_{i \neq j} \Big[\operatorname{Cov}(\mathbf{z}_1)_{ij}^2 + \operatorname{Cov}(\mathbf{z}_2)_{ij}^2\Big].
\end{aligned}
\end{equation}


\noindent\textit{Adversarial domain alignment.}
To encourage modality invariance, we introduce a discriminator $D_\eta:\mathbb{R}^h\!\to\!\{0,1\}$ that predicts whether an embedding comes from ST or SC. In our implementation, the domain loss is computed on non-augmented embeddings: $\mathbf{z}=f_\theta(\mathbf{x})$. Let $\mathcal{L}_{\text{adv}}$ be the discriminator cross-entropy. We minimize $\mathcal{L}_{\text{adv}}$ w.r.t.\ $\eta$, and update $f_\theta$ through a GRL that multiplies discriminator gradients by $-\alpha_{\text{GRL}}$ before they reach $\theta$:
\begin{equation}
\min_{\theta}\;\max_{\eta}\;
\mathcal{L}_{\text{VICReg}}
-\lambda_{\text{adv}}\,\mathcal{L}_{\text{adv}}\bigl(D_\eta,f_\theta\bigr).
\end{equation}
After convergence, we freeze $f_\theta$ and use the resulting embeddings for all subsequent stages.

\vspace{-0.25cm}
\subsection{Minisets for Geometric Supervision}
\textbf{Why construct local training sets rather than use full slides?}
ST being an expensive process, datasets usually contain few slides, whereas diffusion models require many training examples. We increase the effective sample size by extracting many overlapping local subsets (\emph{minisets}) per slide, which preserves informative neighborhood geometry while improving data efficiency.

\noindent\textbf{\ul{\emph{Spatially localized sampling.}}}
For each miniset, we sample a size $n \in [n_{\min}, n_{\max}]$ and a center spot $c$, then draw the remaining $n-1$ spots from a local candidate pool with probability
$p(i\mid c) \propto \exp(-\delta_{i,c}/\tau_{\text{spatial}})$,
where $\delta_{i,c}$ is Euclidean distance in ST coordinate space. This yields spatially coherent minisets while maintaining diversity across draws.

\noindent\textbf{\ul{\emph{Paired overlapping minisets.}}}
Each training example is a pair of minisets from the same slide with a controlled shared index set $\mathcal{I}$ of size $|\mathcal{I}|=\max(n_{\min}^{\text{overlap}},\lfloor \alpha n \rfloor)$. We include $\mathcal{I}$ in both views and sample the remaining points independently from the same (or a nearby) center, so shared spots appear under different contexts for overlap-consistency losses.

\noindent\textbf{\ul{\emph{Pose-invariant geometric supervision.}}}
Absolute ST coordinates encode arbitrary imaging choices (orientation and global positioning) that should not influence learned geometry, so we supervise only intrinsic properties invariant to rigid transformations.
For a miniset with coordinates $\mathbf{P}\in\mathbb{R}^{n\times 2}$, we center to remove translation ($\mathbf{Y}=\mathbf{P}-\bar{\mathbf{P}}$) and compute the Gram matrix $\mathbf{G}=\mathbf{Y}\mathbf{Y}^\top\in\mathbb{R}^{n\times n}$.
We obtain a canonical factor by eigendecomposition $\mathbf{G}=\mathbf{U}\mathbf{\Lambda}\mathbf{U}^\top$ and define
$\mathbf{V}_{\text{target}}=\mathbf{U}_{:,1:d}\big(\mathbf{\Lambda}_{1:d,1:d}\big)^{1/2}\in\mathbb{R}^{n\times d}$, so that $\mathbf{V}_{\text{target}}\mathbf{V}_{\text{target}}^\top=\mathbf{G}$.
We use an overcomplete $d=32$; for planar sections $\mathbf{G}$ is low-rank, and Gram-based losses emphasize the intrinsic low-dimensional structure. The factorization is not unique: $\mathbf{V}_{\text{target}}\mathbf{Q}$ yields the same $\mathbf{G}$ for any $\mathbf{Q}\in O(d)$, where $O(d)$ is the group of $d\times d$ orthogonal matrices ($\mathbf{Q}^\top\mathbf{Q}=\mathbf{I}$). Overall, the model is supervised to match intrinsic geometry through $\mathbf{G}$, rather than absolute positions.

\vspace{-0.5cm}
\subsection{Conditional Geometry Generation}
\noindent\textbf{Why frame spatial reconstruction as conditional generation?}
ST-derived targets specify \emph{intrinsic} tissue geometry but do not define a deterministic mapping from gene expression to spatial location. Dissociated SC profiles arrive without coordinates and without an assignment to a measured ST lattice. We therefore learn a conditional generative model that, given an expression-conditioned set, produces a continuous geometric configuration whose intrinsic structure matches that observed in ST.

\noindent\textbf{\ul{\emph{Coordinate-free geometry generation.}}} Given a miniset of expression embeddings $\{f_\theta(\mathbf{x}_i)\}_{i=1}^n$, the model predicts a latent geometry $\mathbf{V}\in\mathbb{R}^{n\times d}$. Supervision is applied through intrinsic functions of $\mathbf{V}$, including its Gram matrix $\mathbf{V}\mathbf{V}^\top$ and pairwise distances, which are invariant to global orthogonal transforms: for any $\mathbf{Q}\in O(d)$, $(\mathbf{V}\mathbf{Q})(\mathbf{V}\mathbf{Q})^\top=\mathbf{V}\mathbf{V}^\top$. For stable diffusion training, we fix a deterministic canonical gauge by aligning $\mathbf{V}$ to the canonical target factorization $\mathbf{V}_{\text{target}}$ from Sec.~3.2, which preserves intrinsic geometry while removing arbitrary pose.

\vspace{-0.5cm}
\subsection{Geometry-Aware Diffusion Training}

\noindent\textbf{Why use permutation-equivariant architectures?} Minisets are unordered sets of spots sampled stochastically from tissue. The model must therefore be permutation-equivariant: permuting spot indices should permute outputs in the same way, without changing predicted spatial relationships. We use a Set Transformer~\cite{lee2019set} backbone with induced set attention blocks (ISAB), which supports variable-sized sets and captures global context while remaining permutation-equivariant.

\noindent\textbf{\ul{\emph{Context encoder.}}} For a miniset, let $\mathbf{Z}_{\text{set}}\in\mathbb{R}^{n\times h}$ denote the expression embeddings produced by the shared encoder $f_\theta$ trained in the previous subsection (and frozen thereafter). The context encoder $f_\phi$ maps these to contextual features $\mathbf{H}=f_\phi(\mathbf{Z}_{\text{set}})\in\mathbb{R}^{n\times c}$, where $c$ is the context feature dimension (a hyperparameter).
Concretely, $f_\phi$ applies a linear projection followed by stacked ISAB layers; using $m$ inducing points (a small latent set that mediates attention between elements) reduces attention complexity from $O(n^2)$ to $O(nm)$.

\noindent\textbf{\ul{\emph{Generator for coarse geometry proposals.}}} A generator $\Phi_\psi$ maps context features to an initial geometric proposal
$\mathbf{V}_{\text{base}}=\Phi_\psi(\mathbf{H})\in\mathbb{R}^{n\times d}$.
We implement $\Phi_\psi$ as additional ISAB layers followed by an MLP head, and mean-center $\mathbf{V}_{\text{base}}$ across spots to remove arbitrary translations. This proposal provides a coarse, expression-consistent configuration and serves as the starting point for diffusion-based refinement; $\Phi_\psi$ is trained jointly with the diffusion model under the same geometric supervision.

\noindent\textbf{\ul{\emph{EDM-preconditioned score network.}}}
We refine geometry with an EDM-preconditioned denoiser~\cite{karras2022elucidatingdesignspacediffusionbased}. Given a noisy input $\mathbf{U}_t\in\{\mathbf{V}_t,\mathbf{R}_t\}$ at noise level $\sigma$, context features $\mathbf{H}$, and noise embedding $c_{\text{noise}}(\sigma)$ (with self-conditioning), a permutation-equivariant ISAB backbone predicts a clean estimate via
\begin{equation}
\Psi_\omega(\mathbf{U}_t,\sigma)
=
c_{\text{skip}}(\sigma)\,\mathbf{U}_t
+
c_{\text{out}}(\sigma)\,
\mathbf{F}_\omega\!\big(c_{\text{in}}(\sigma)\,\mathbf{U}_t;\,c_{\text{noise}}(\sigma),\mathbf{H}\big),
\end{equation}
where $c_{\text{skip}}(\sigma)=\frac{\sigma_{\text{data}}^2}{\sigma^2+\sigma_{\text{data}}^2}$,
$c_{\text{out}}(\sigma)=\frac{\sigma\sigma_{\text{data}}}{\sqrt{\sigma^2+\sigma_{\text{data}}^2}}$,
$c_{\text{in}}(\sigma)=\frac{1}{\sqrt{\sigma^2+\sigma_{\text{data}}^2}}$,
and $c_{\text{noise}}(\sigma)=\tfrac14\log\sigma$; $\sigma_{\text{data}}$ is the empirical scale of the clean targets ($\sigma_{\text{data,resid}}$ in residual mode, defined below).
We mean-center over valid spots before Gram-based losses so translation does not affect supervision.

\noindent\textbf{\ul{\emph{Residual diffusion mode.}}} Rather than denoising from pure noise to final geometry, we factorize the generation process: the generator produces $\mathbf{V}_{\text{base}}$ encoding coarse structure, while the diffusion model denoises residual corrections. For each training sample, we apply orthogonal alignment (rotations and reflections, no scaling) to align the target geometry to the generator's frame: $\mathbf{V}_{\text{target,aligned}} = \text{Align}(\mathbf{V}_{\text{target}}, \mathbf{V}_{\text{base}})$ using Procrustes analysis with transformations in $O(d)$(after mean-centering). The residual target is $\mathbf{R}_{\text{target}} = \mathbf{V}_{\text{target,aligned}} - \mathbf{V}_{\text{base}}$. We add noise to this residual: $\mathbf{R}_t = \mathbf{R}_{\text{target}} + \sigma \epsilon$ where $\epsilon \sim \mathcal{N}(0, \mathbf{I})$, and the score network predicts the clean residual $\hat{\mathbf{R}}_0=\Psi_\omega(\mathbf{R}_t,\sigma)$. Final predictions compose: $\mathbf{V}_{\text{pred}} = \mathbf{V}_{\text{base}} + \hat{\mathbf{R}}_0$. The key advantage is scale: aligned residuals have substantially smaller magnitude than absolute coordinates ($\sigma_{\text{data,resid}} \ll \sigma_{\text{data}}$). EDM coefficients use $\sigma_{\text{data,resid}}$, ensuring the network operates in a properly normalized range and yielding a better-conditioned denoising problem.

\noindent\textbf{\ul{\emph{Curriculum scheduling for progressive noise coverage.}}}
We train with a stage-wise noise cap $\sigma_{\text{cap}}$ that increases over training: early stages sample $\sigma\in[\sigma_{\min},\sigma_{\text{cap}}]$ at moderate noise where geometric structure remains visible, and later stages extend $\sigma_{\text{cap}}$ toward $\sigma_{\max}$. At each stage, we sample noise levels on a log scale (stratified for coverage) to avoid under-training high-$\sigma$ regions. Residual diffusion makes curriculum more tractable because the same $\sigma$ obscures less information when applied to residuals than to absolute geometry.

\noindent\textbf{\ul{\emph{Training objectives.}}}
We train with an EDM-weighted denoising loss plus auxiliary geometry losses that directly penalize structural errors. In residual mode, the clean target is $\mathbf{R}_{\text{target}}$ (defined above). We sample $\boldsymbol{\epsilon}\sim\mathcal{N}(\mathbf{0},\mathbf{I})$ and form $\mathbf{R}_t=\mathbf{R}_{\text{target}}+\sigma\boldsymbol{\epsilon}$.  With a validity mask $\mathbf{M}\in\{0,1\}^{n\times 1}$ (broadcast across the $d$ geometry dimensions), the primary objective is
\begin{equation}
\mathcal{L}_{\text{score}}
=
\mathbb{E}_{\sigma,\boldsymbol{\epsilon}}
\left[
w(\sigma)\,
\left\|
\big(\hat{\mathbf{R}}_0-\mathbf{R}_{\text{target}}\big)\odot \mathbf{M}
\right\|_F^2
\right],
\qquad
\hat{\mathbf{R}}_0=\Psi_\omega(\mathbf{R}_t,\sigma),
\end{equation}
where $w(\sigma) = (\sigma^2 + \sigma_{\text{data,resid}}^2)/(\sigma \cdot \sigma_{\text{data,resid}})^2$ is the standard EDM weighting in residual space.

\noindent\textbf{\ul{\emph{Gram-based geometry losses.}}}
The denoising loss trains $\Psi_\omega$ to predict $\mathbf{R}_{\text{target}}$, but we additionally enforce that the composed geometry
$\mathbf{V}_{\text{pred}}=\mathbf{V}_{\text{base}}+\hat{\mathbf{R}}_0$
matches the intrinsic target structure. Let
$\mathbf{G}_{\text{pred}}=\mathbf{V}_{\text{pred}}\mathbf{V}_{\text{pred}}^\top$
and
$\mathbf{G}_{\text{target}}=\mathbf{V}_{\text{target,aligned}}\mathbf{V}_{\text{target,aligned}}^\top$.
We use a scale-normalized Gram loss
\begin{equation}
\mathcal{L}_{\text{gram}}
=
\frac{\left\|\mathbf{G}_{\text{pred}}-\mathbf{G}_{\text{target}}\right\|_F^2}
{\left\|\mathbf{G}_{\text{target}}\right\|_F^2},
\end{equation}
and a global scale matching term via log-trace,
\begin{equation}
\mathcal{L}_{\text{gram,scale}}
=
\left(
\log \operatorname{tr}(\mathbf{G}_{\text{pred}})
-
\log \operatorname{tr}(\mathbf{G}_{\text{target}})
\right)^2.
\end{equation}
We apply Gram losses only at sufficiently low noise (high SNR), where geometric structure is observable.

\noindent\textbf{\ul{\emph{Local neighborhood preservation.}}}
To preserve $k$NN structure beyond global Gram matching, we compute target neighborhoods $\mathcal{N}_k(i)$ from $\mathbf{V}_{\text{target,aligned}}$ and apply an NCA-style loss on the predicted geometry:

\begin{equation}
\mathcal{L}_{\text{NCA}}
=
-\sum_{i=1}^n
\log
\frac{
\sum_{j\in \mathcal{N}_k(i)}
\exp\!\left(-\|\mathbf{v}_i^{\text{pred}}-\mathbf{v}_j^{\text{pred}}\|_2^2/\tau_{\text{NCA}}\right)
}{
\sum_{j\neq i}
\exp\!\left(-\|\mathbf{v}_i^{\text{pred}}-\mathbf{v}_j^{\text{pred}}\|_2^2/\tau_{\text{NCA}}\right)
},
\end{equation}
where $\mathbf{v}_i^{\text{pred}}$ is row $i$ of $\mathbf{V}_{\text{pred}}$. In addition, we include an edge-wise local scale penalty on $k$NN edges that penalizes discrepancies in log neighbor distances between $\mathbf{V}_{\text{pred}}$ and $\mathbf{V}_{\text{target,aligned}}$.

\noindent\textbf{\ul{\emph{Generator supervision.}}}
We train the generator output $\mathbf{V}_{\text{base}}$ with direct geometric supervision so that diffusion refines a strong proposal. We use a Procrustes-aligned regression loss:
\begin{equation}
\mathcal{L}_{\text{gen,align}}
=
\min_{\mathbf{Q}\in O(d)}
\left\|
\mathbf{V}_{\text{base}}\mathbf{Q}-\mathbf{V}_{\text{target,aligned}}
\right\|_F^2,
\end{equation}
and a Gram loss:
\begin{equation}
\mathcal{L}_{\text{gen,gram}}
=
\left\|
\mathbf{V}_{\text{base}}\mathbf{V}_{\text{base}}^\top
-
\mathbf{G}_{\text{target}}
\right\|_F^2,
\end{equation}
optionally augmented with a global scale matching term (e.g., log-RMS) to ensure the proposal magnitude is calibrated.

\noindent\textbf{\ul{\emph{Overlap consistency for context-invariant learning.}}}
We sample paired overlapping minisets throughout training, and enforce that shared points receive consistent geometric predictions even when their surrounding context differs. Let $\mathcal{I}$ be the shared index set and let $\mathbf{V}_1^{\mathcal{I}},\mathbf{V}_2^{\mathcal{I}}\in\mathbb{R}^{|\mathcal{I}|\times d}$ denote the predicted geometries restricted to $\mathcal{I}$ from the two views. After mean-centering over $\mathcal{I}$, we form Gram matrices
$\mathbf{G}_1=\mathbf{V}_1^{\mathcal{I}}(\mathbf{V}_1^{\mathcal{I}})^\top$
and
$\mathbf{G}_2=\mathbf{V}_2^{\mathcal{I}}(\mathbf{V}_2^{\mathcal{I}})^\top$,
and penalize normalized shape disagreement:
\begin{equation}
\mathcal{L}_{\text{ov,shape}}
=
\left\|
\frac{\mathbf{G}_1}{\operatorname{tr}(\mathbf{G}_1)}
-
\frac{\mathbf{G}_2}{\operatorname{tr}(\mathbf{G}_2)}
\right\|_F^2.
\end{equation}
We optionally add a scale term
$\mathcal{L}_{\text{ov,scale}}=\big(\log\operatorname{tr}(\mathbf{G}_1)-\log\operatorname{tr}(\mathbf{G}_2)\big)^2$
and a neighborhood-consistency term via symmetric KL divergence between softmax distance distributions on $\mathcal{I}$. Overlap losses are applied only when the denoising SNR is sufficiently high for structure to be meaningful (and are disabled at very high noise).

\noindent\textbf{\ul{\emph{Joint training.}}}
We train the context encoder $f_\phi$, generator $\Phi_\psi$, and denoiser $\Psi_\omega$ jointly with a weighted sum of the objectives above (denoising, Gram and $k$NN structure, generator supervision, and overlap consistency). We use standard stabilizers (learning-rate scheduling, gradient clipping).

\vspace{-0.3cm}
\subsection{Patchwise Single-Cell Reconstruction}
\label{sec:inference}

\noindent\textbf{Why patchwise inference?}
The geometry model is trained on minisets, whereas single-cell datasets typically contain many more cells than a single ST slide. One-shot generation on all $N_{SC}$ cells is therefore (i) computationally prohibitive and (ii) mismatched to the miniset training regime. We instead sample many \emph{overlapping patches} of cells, predict \emph{local} geometries, convert them into \emph{distance measurements}, and aggregate these distances into a robust global distance graph (and, optionally, a dense Euclidean distance matrix via Step~7).

\paragraph{Step 1: Encode all cells and build a locality graph.}
Let $\mathbf{E}_{\text{SC}}\in\mathbb{R}^{N_{SC}\times G}$ be single-cell expression and compute shared embeddings $\mathbf{Z}_{\text{all}}=f_\theta(\mathbf{E}_{\text{SC}})\in\mathbb{R}^{N_{SC}\times h}$ using the frozen encoder. We construct a mutual-$k$NN graph in $\mathbf{Z}_{\text{all}}$ space and retain only robust edges using Jaccard overlap filtering: an edge $(i,j)$ is kept only if $i\in\mathcal{N}^Z_k(j)$ and $j\in\mathcal{N}^Z_k(i)$ and the Jaccard score $J(i,j)=\frac{|\mathcal{N}^Z_k(i)\cap \mathcal{N}^Z_k(j)|}{|\mathcal{N}^Z_k(i)\cup \mathcal{N}^Z_k(j)|}$ exceeds a threshold $\tau_{J}$. This yields a locality graph $\mathcal{G}_Z=(\{1,\dots,N_{SC}\},\mathcal{E}_Z)$ used only for patch sampling.

\paragraph{Step 2: Sample overlapping patches by random walks.}
We sample a collection of index sets (patches) $\{\mathcal{S}_p\}_{p=1}^{N_{\text{patches}}}$ of fixed size $|\mathcal{S}_p|=n_{\text{patch}}$ using random walks on $\mathcal{G}_Z$. We enforce a minimum overlap $|\mathcal{S}_p\cap \mathcal{S}_q|\ge n_{\min}^{\text{overlap}}$ for neighboring patches so that the union of patches forms a connected cover of cells.

\paragraph{Step 3: Per-patch geometry generation.}
For each patch $\mathcal{S}_p$, we extract $\mathbf{Z}_{\text{set}}^{(p)}=\mathbf{Z}_{\text{all}}[\mathcal{S}_p]\in\mathbb{R}^{n_{\text{patch}}\times h}$, compute context $\mathbf{H}^{(p)}=f_\phi(\mathbf{Z}_{\text{set}}^{(p)})$, and obtain a coarse proposal $\mathbf{V}_{\text{base}}^{(p)}=\Phi_\psi(\mathbf{H}^{(p)})\in\mathbb{R}^{n_{\text{patch}}\times d}$. Residual diffusion then refines this proposal under a fixed noise schedule $\{\sigma_\ell\}_{\ell=1}^L$, producing $\mathbf{V}_{\text{pred}}^{(p)}\in\mathbb{R}^{n_{\text{patch}}\times d}$.

\paragraph{Step 4: Convert patch geometries into distance measurements.}
From each $\mathbf{V}_{\text{pred}}^{(p)}$, we extract local distance measurements on a chosen within-patch edge set $\mathcal{E}^{(p)}\subseteq\mathcal{S}_p\times\mathcal{S}_p$ (e.g., $k$NN edges in the patch geometry). For $(i,j)\in\mathcal{E}^{(p)}$, the measured distance is $\hat{d}_{ij}^{(p)}=\|\mathbf{v}_i^{(p)}-\mathbf{v}_j^{(p)}\|_2$, where $\mathbf{v}_i^{(p)}$ denotes the row of $\mathbf{V}_{\text{pred}}^{(p)}$ corresponding to global cell index $i\in\mathcal{S}_p$.


\paragraph{Step 5: Overlap consistency and patch reliability.}
For overlapping patches $p,q$, let $\mathcal{I}_{pq}=\mathcal{S}_p\cap\mathcal{S}_q$ be shared cells (pairs with $|\mathcal{I}_{pq}|<5$ are skipped). We quantify disagreement by the mean absolute difference between \emph{log} pairwise distances on the overlap (computed from the patch geometries). Each patch receives a reliability weight $a_p\in(0,1]$ by exponentiating its mean disagreement with neighboring patches, normalized by the median disagreement across all overlap pairs, so patches that disagree more are down-weighted.

\paragraph{Step 6: Aggregate a global distance graph.}
For any global pair $(i,j)$ observed in multiple patches, let $\mathcal{P}_{ij}$ denote the set of patches that measured $(i,j)$.
We aggregate the distances $\{\hat d_{ij}^{(p)}\}_{p\in\mathcal{P}_{ij}}$ using a reliability-weighted \emph{median} (with weights $a_p$), which is robust to outlier patches. We keep only edges with sufficient support and low dispersion across patches (minimum count and a relative spread threshold), and assign each retained edge a confidence weight $\omega_{ij}$ that increases with measurement count and decreases with dispersion. This yields a stitched distance graph $\mathcal{G}_D=(\{1,\dots,N_{SC}\},\mathcal{E}_D,\hat{\mathbf{d}},\boldsymbol{\omega})$.

\paragraph{Step 7: Global 2D distance-geometry solve and output.}
We initialize $\mathbf{X}\in\mathbb{R}^{N_{SC}\times 2}$ using Landmark Isomap and refine it by minimizing a weighted \emph{Huber} distance-geometry objective on $\mathcal{G}_D$, with a small anchor term to keep $\mathbf{X}$ close to the initialization. During optimization we monitor stress and edge residual statistics (mean/median/max) as empirical checks of (near-)Euclidean consistency. We return canonical coordinates $\mathbf{X}$ and define the dense distance matrix $\mathbf{D}\in\mathbb{R}^{N_{SC}\times N_{SC}}$ by $D_{ij}=\|\mathbf{x}_i-\mathbf{x}_j\|_2$.

\vspace{-0.3cm}
\begin{table*}[t]
\centering
\captionsetup{font=small}
\caption{Spatial reconstruction benchmark. \textbf{Bold}: best, \underline{underline}: 2nd best, \textit{italic}: 3rd best.}
\label{tab:benchmark}
\renewcommand{\arraystretch}{1.15}
\resizebox{\textwidth}{!}{%
\begin{tabular}{c|l|ccc|ccc|cc|cc}
\hline
\multirow{2}{*}{\textbf{Dataset}} & \multirow{2}{*}{\textbf{Method}} & \multicolumn{3}{c|}{\textbf{Global Geometry}} & \multicolumn{3}{c|}{\textbf{Local Geometry}} & \multicolumn{2}{c|}{\textbf{Neighborhood Quality}} & \multicolumn{2}{c}{\textbf{Distribution}} \\
\cline{3-5} \cline{6-8} \cline{9-10} \cline{11-12}
 &  & \makecell{Spearman\\$(\mathbf{D}, \mathbf{D}^{\mathrm{GT}})$ $\uparrow$} 
 & \makecell{Pearson\\$(\mathbf{D}, \mathbf{D}^{\mathrm{GT}})$ $\uparrow$} 
 & \makecell{Stress-1\\$(\mathbf{D}, \mathbf{D}^{\mathrm{GT}})$ $\downarrow$} 
 & \makecell{Edge ROC-AUC\\$(\mathbf{D}; R_{20})$ $\uparrow$} 
 & \makecell{bAP\\$(\mathbf{D}; R_{20})$ $\uparrow$} 
 & \makecell{Shell F1\\(macro) $\uparrow$} 
 & \makecell{Trust@20 $\uparrow$} 
 & \makecell{Cont@20 $\uparrow$} 
 & \makecell{SWD\\$(\mathbf{X}, \mathbf{X}^{\mathrm{GT}})$ $\downarrow$} 
 & \makecell{$W_1$($k$NN-dist;\\$k$=20) $\downarrow$} \\

\hline
& Tangram & 0.7938 & 0.8040 & 0.2765 & \textit{0.9001} & \underline{0.8929} & \textbf{0.4847} & \underline{0.8979} & \underline{0.9027} & \textbf{0.0393} & \underline{0.0058} \\
& novoSpaRc & 0.0539 & 0.0436 & 0.6639 & 0.8344 & 0.8094 & 0.1319 & \textit{0.8806} & \textit{0.8904} & 0.1558 & 0.0237 \\
 & STEM & \underline{0.8251} & \underline{0.8318} & \textbf{0.1701} & \underline{0.9007} & \textit{0.8929} & \underline{0.4551} & 0.8806 & 0.8904 & \underline{0.0393} & 0.0103 \\
 & SpaOTsc & 0.4778 & 0.5014 & 0.4333 & 0.6341 & 0.6009 & 0.1776 & 0.7112 & 0.7294 & 0.0598 & 0.0118 \\
 & cell2location & 0.1758 & 0.1597 & 0.7258 & 0.7729 & 0.7624 & 0.0744 & 0.6836 & 0.7720 & 0.3633 & 0.0368 \\
 & scSpace & 0.0045 & 0.0054 & 0.9913 & 0.5307 & 0.5510 & 0.0384 & 0.5154 & 0.5419 & 0.5854 & 0.0382 \\
 & CytoSPACE & \textit{0.8198} & \textit{0.8118} & \textit{0.2471} & 0.8805 & 0.8612 & 0.3984 & 0.8501 & 0.8517 & 0.0949 & 0.0111 \\
 & CeLEry & 0.5729 & 0.5913 & 0.3979 & 0.7641 & 0.7370 & 0.1871 & 0.7677 & 0.7694 & 0.0526 & \textit{0.0095} \\
 & COME & 0.2208 & 0.1116 & 0.7413 & 0.6453 & 0.6575 & 0.1220 & 0.6466 & 0.6811 & 0.2021 & 0.0231 \\
\multirow{-10}{*}{\rotatebox[origin=c]{90}{\large\textbf{\makecell{Mouse Atlas\\(seqFISH$\rightarrow$pseudo-Visium)}}}} & GEARS & \textbf{0.8324} & \textbf{0.8331} & \underline{0.1979} & \textbf{0.9327} & \textbf{0.9063} & \textit{0.4279} & \textbf{0.8998} & \textbf{0.9155} & \textit{0.0407} & \textbf{0.0026} \\
\hline
 & Tangram & 0.2901 & 0.2901 & 0.5341 & 0.6871 & 0.6775 & 0.1471 & 0.6755 & 0.7284 & 0.1361 & 0.0645 \\
 & novoSpaRc & 0.1651 & 0.1603 & 0.636 & 0.6677 & 0.6387 & 0.1253 & 0.5933 & 0.6950 & 0.2709 & 0.1619 \\
 & STEM & 0.2945 & 0.2859 & 0.5715 & 0.7121 & \textbf{0.7846} & 0.1368 & 0.6539 & 0.7530 & 0.2188 & 0.0997 \\
 & SpaOTsc & 0.3086 & 0.2918 & 0.5513 & 0.7350 & 0.7194 & 0.1442 & 0.7421 & \textit{0.7667} & 0.1918 & 0.0801 \\
 & cell2location & — & — & — & — & — & — & — & — & — & — \\
 & scSpace & \underline{0.5421} & \underline{0.5339} & \underline{0.4208} & \textbf{0.7952} & \underline{0.7647} & \underline{0.1862} & \textbf{0.7701} & \textbf{0.8203} & 0.0880 & 0.0475 \\
 & CytoSPACE & 0.3488 & 0.3545 & 0.4771 & 0.6755 & 0.6673 & 0.1668 & 0.6850 & 0.6850 & \underline{0.0440} & \underline{0.0260} \\
 & CeLEry & \textit{0.4745} & \textit{0.4797} & \textit{0.4325} & \textit{0.7516} & 0.7215 & \textit{0.1729} & \textit{0.7423} & 0.7643 & \textit{0.0550} & \textit{0.0359} \\
 & COME & 0.2441 & 0.0517 & 0.8966 & 0.6939 & 0.7031 & 0.0528 & 0.6790 & 0.7575 & 0.3455 & 0.1577 \\
\multirow{-10}{*}{\rotatebox[origin=c]{90}{\large\textbf{\makecell{hSCC\\(multi-slide Visium)}}}} & GEARS & \textbf{0.5468} & \textbf{0.5456} & \textbf{0.4040} & \underline{0.7874} & \textit{0.7612} & \textbf{0.2126} & \underline{0.7520} & \underline{0.7934} & \textbf{0.0389} & \textbf{0.0256} \\
\hline
\end{tabular}%
}
\vspace{-1mm}
\noindent\rule{\textwidth}{0.3pt}
\vspace{-0.5mm}
{\small \textbf{Note:} $\mathbf{D}^{\mathrm{GT}}$ = Ground truth (GT) pairwise distances, $\mathbf{D}$ = predicted pairwise distances, $\mathbf{X}^{\mathrm{GT}}/\mathbf{X}$ = GT/pred coordinates, $R_{20}$ = median 20-NN radius from GT. cell2location results on hSCC are omitted due to missing cell-type annotations required for its reference signatures.}
\end{table*}

\section{Experiment}
\subsection{Experimental Setup}
\noindent\textbf{Dataset Description.}
We evaluate on two datasets. \textbf{(i) Spatial Mouse Atlas (seqFISH)}~\cite{lohoff2020highly}. We use a single-cell--resolution, highly multiplexed in situ ST mouse embryo atlas (seqFISH) that provides per-cell coordinates together with a targeted gene panel. Following the standard benchmarking protocol~\cite{hao2024stem}, we derive a Visium-like spot layer by overlaying a regular grid and aggregating nearby cells into pseudo-spots (discarding low-coverage spots); the original cell coordinates are retained as ground truth for quantitative evaluation of reconstructed single-cell geometry. (ii) \textbf{Human squamous cell carcinoma (hSCC).} We use a multi-section hSCC dataset with spot-based ST slides and a paired scRNA-seq cohort. We train the model on two ST slides and evaluate generalization on a held-out third slide by running inference on its spot expression and comparing the reconstructed intrinsic geometry to the slide’s measured coordinates (pose-invariant metrics).


\noindent\textbf{Baseline Algorithms.}
We compared against nine representative baselines spanning both \emph{cell-to-space alignment} and \emph{spot-level deconvolution} paradigms:
1. \textbf{Tangram}~\cite{biancalani2021deep} learns a soft cell-to-spot mapping matrix by aligning shared-gene expression between cells and spots, yielding per-cell placement via expected spot coordinates;
2. \textbf{novoSpaRc}~\cite{moriel2021novosparc} reconstructs spatial organization by optimizing probabilistic assignments of cells to spatial locations such that reconstructed spatial gene expression matches measured patterns;
3. \textbf{STEM}~\cite{hao2024stem} learns spatially informed embeddings and uses ST spatial adjacency graph to supervise alignment, producing a cell-to-spot mapping matrix and induced 2D placements;
4. \textbf{SpaOTsc}~\cite{cang2020inferring} formulates mapping as structured optimal transport between cells and spatial location, returning a coupling matrix that induces cell placements via weighted spot coordinates.
5. \textbf{cell2location}~\cite{kleshchevnikov2022cell2location} is a spot-level deconvolution model that infers spot-wise abundances of reference-derived cell states from SC and ST expression;
6. \textbf{scSpace}~\cite{qian2023reconstruction} is a coordinate-regression method that learns a mapping from SC expression to spatial coordinates by training on ST spots with known positions, enabling direct point estimates of cell locations in the plane.
7. \textbf{CytoSPACE}~\cite{vahid2023high} performs constrained cell placement onto spots (e.g., enforcing per-spot capacity and consistency with cell-type composition constraints) to obtain near-discrete cell-to-spot assignments;
8. \textbf{CeLEry}~\cite{zhang2023leveraging} trains a supervised coordinate-regression model that predicts ell coordinates directly from SC expression using ST coordinates as supervision;
9. \textbf{COME}~\cite{wei2025come} learns a cell-to-spot probability matrix via contrastive mapping objectives between SC and ST profiles, enabling per-cell placement through the inferred spot-weight distribution.

\vspace{-0.4cm}
\subsection{Experimental Results}
\label{sec:overall_results}
\noindent\textbf{Overall comparison}. Table~\ref{tab:benchmark} summarizes reconstruction quality across four aspects: (i) \emph{global geometry} agreement between predicted and ground-truth pairwise distances, (ii) \emph{local geometry} and edge recovery at the neighborhood scale, (iii) \emph{neighborhood quality} (trustworthiness/continuity), and (iv) \emph{distribution matching} between reconstructed and ground-truth spatial layouts. For methods that output coordinates, we form predicted distances $\mathbf{D}$ by Euclidean distances between predicted points; for distance-first methods (including ours), $\mathbf{D}$ is produced directly by the model and/or by the distance-geometry solve. Evaluation metric details in Appendix~\ref{app:metrics}.

\noindent\textbf{Global geometry.}
Global geometry measures whether the reconstruction preserves \emph{overall} pairwise structure: Spearman/Pearson correlations quantify rank and linear agreement between predicted and ground-truth (GT) distances, while Stress-1 reports normalized global distance distortion. On Mouse Atlas, our method achieves the strongest global agreement(Spearman/Pearson) while remaining competitive on Stress-1 (Table~\ref{tab:benchmark}), outperforming mapping-based baselines such as Tangram, novoSpaRc, SpaOTsc, and STEM. On the hSCC generalization setting (train two ST slides, evaluate on a held-out slide), our method yields the best global structure overall, indicating that the reconstructed intrinsic geometry transfers across tissue sections.

\noindent\textbf{Local geometry.}
Local geometry evaluates whether \emph{nearby} relations are preserved: Edge ROC-AUC treats GT near/far pairs as a binary task using predicted distances as scores, balanced average precision (bAP) summarizes precision-recall under class imbalance, and Shell-F1 evaluates distance preservation across multiple local shells (multi-scale). On Mouse Atlas, our method achieves the best edge recovery and balanced precision, reflecting accurate local adjacency compared to Tangram/ novoSpaRc/ SpaOTsc/ STEM and coordinate-regression baselines (CeLEry, scSpace). On hSCC, we remain competitive on Edge ROC-AUC while attaining the best Shell-F1, suggesting improved multi-scale local structure even when a baseline may win a single metric.

\begin{figure*}[t]
    \centering
    \includegraphics[width=\textwidth]{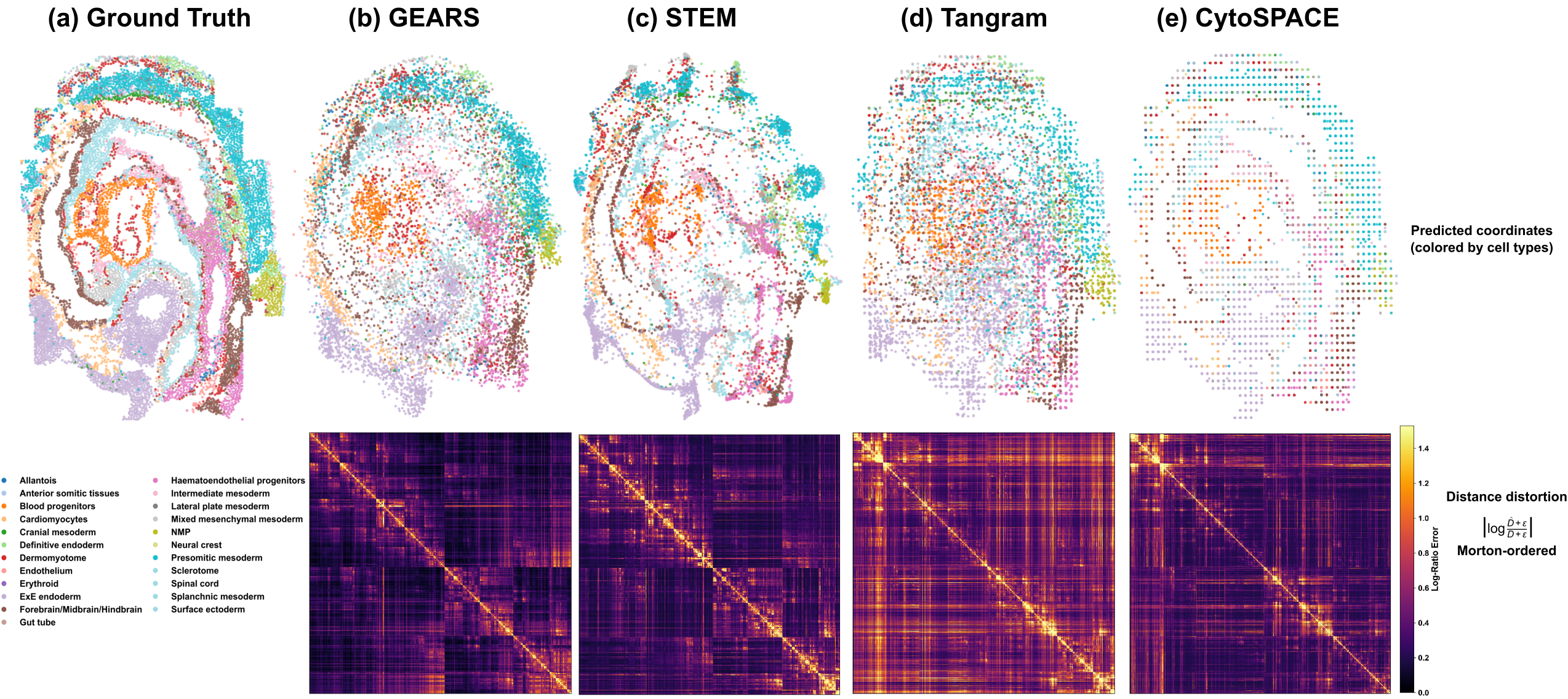}
    \captionsetup{justification=centering, font=small}
    \caption{\textbf{Qualitative reconstruction and distance-distortion diagnostics on Mouse Atlas.}
    Top: ground-truth cell coordinates (a) and predicted coordinates from GEARS (ours) (b) and representative baselines STEM (c), Tangram (d), and CytoSPACE (e).
    Bottom: pairwise distance distortion heatmaps for each method, defined as $E_{ij}=|\log((\hat D_{ij}+\epsilon)/(D_{ij}+\epsilon))|$, where $D_{ij}$ and $\hat D_{ij}$ are ground-truth and predicted pairwise distances after optimal global scaling.
    Cells are ordered by a Morton (space-filling curve) ordering of the ground-truth coordinates to preserve spatial locality; heatmaps are block-averaged for visualization.
    Brighter values indicate larger multiplicative distortion of inter-cell distances.}
    \vspace{-0.3cm}
    \label{fig:qual_mouseatlas}
\end{figure*}

\noindent\textbf{Neighborhood quality.}
Trustworthiness@20 (Trust@20) and Continuity@20 (Cont@20) measure $k$NN($k=20$) fidelity from two directions: Trust@20 penalizes spurious predicted neighbors, while Cont@20 penalizes missing true neighbors. On the Mouse Atlas, our method performs the best on both metrics, indicating stable neighborhood preservation. On hSCC, the top neighborhood-quality scores are achieved by coordinate-regression methods: scSpace ranks first and CeLEry third while GEARS places second; importantly, GEARS attains this neighborhood fidelity alongside stronger global distance agreement.

\noindent\textbf{Distribution matching.}
Distribution metrics compare \emph{overall spatial layout statistics} beyond pairwise correspondence: Sliced Wasserstein Distance (SWD) compares point-cloud shape via random 1D projections (no explicit correspondence), and W$_1$($k$NN-dist; $k=20$) compares the distributions of local neighbor distances (length-scale calibration). On hSCC, our method achieves the best W$_1$, indicating well-calibrated local length scales. On Mouse Atlas, our method attains the best W$_1$ and competitive SWD, suggesting that the reconstructed layout matches the held-out slide’s global shape while preserving local spacing statistics.

\noindent\textbf{Ablation: residual diffusion refinement.}
Table~\ref{tab:ablation} isolates the contribution of the residual diffusion refiner by comparing generator-only inference (w/o Diff) to the full pipeline (Full). Residual diffusion consistently improves distance-geometry fidelity: Stress-1 drops substantially and distance correlations increase. The largest gains are in \emph{scale-sensitive} and \emph{distributional} metrics, with sharp reductions in Scale Err and SWD, indicating that refinement calibrates global distance magnitudes and yields a point-cloud distribution closer to the reference. Improvements on neighborhood-boundary metrics are smaller and occasionally neutral (Shell F1 / Edge ROC on hSCC), consistent with these metrics being sensitive to local ties and near-boundary ambiguity. Overall, the refiner mainly corrects global and meso-scale distortions left by the generator, matching its intended role as a geometry-calibration stage.

\vspace{-0.05cm}
\noindent\textbf{Qualitative geometry recovery and distortion localization.}
Fig.~\ref{fig:qual_mouseatlas} visualizes reconstructions on the Mouse Atlas, where ground-truth single-cell coordinates are available. GEARS recovers a coherent tissue layout that preserves large-scale organization while maintaining separation between major regions (Fig.~\ref{fig:qual_mouseatlas}b). In contrast, STEM produces sharp-looking structure but exhibits \emph{density collapse}, mapping many cells into an unrealistically compact area, which is reflected by elevated off-diagonal distortion in the heatmap (Fig.~\ref{fig:qual_mouseatlas}c). Tangram recovers parts of the global outline but blurs regional structure and shows stronger long-range distortions (Fig.~\ref{fig:qual_mouseatlas}d). CytoSPACE yields a discretized, lattice-like placement with quantization artifacts, visible as structured block patterns in the distortion map (Fig.~\ref{fig:qual_mouseatlas}e). Overall, the distortion heatmaps localize \emph{where} errors occur across scales and support the quantitative improvements reported in Table~\ref{tab:benchmark}.

\begin{table}[]
\centering
\vspace{-0.1cm}
\captionsetup{font=small}
\caption{Ablation study: Effect of residual diffusion refinement. Arrows indicate improvement magnitude.}
\label{tab:ablation}
\vspace{-0.3cm}
\resizebox{\columnwidth}{!}{%
\begin{tabular}{l|ccc|ccc}
\hline
\multirow{2}{*}{\textbf{Metric}} & \multicolumn{3}{c|}{\textbf{Mouse Atlas}} & \multicolumn{3}{c}{\textbf{hSCC}} \\
\cline{2-7}
 & w/o Diff & Full & $\Delta$\% & w/o Diff & Full & $\Delta$\% \\
\hline
Stress-1 $\downarrow$ & 0.3007 & \textbf{0.1979} & \textcolor{green!50!black}{+34.2 $\uparrow\uparrow$} & 0.6079 & \textbf{0.4040} & \textcolor{green!50!black}{+33.5 $\uparrow\uparrow$} \\
Scale Err $\downarrow$ & 5.9600 & \textbf{1.3825} & \textcolor{green!50!black}{+76.8 $\uparrow\uparrow$} & 15.0300 & \textbf{12.7000} & \textcolor{green!50!black}{+15.5 $\uparrow$} \\
SWD $\downarrow$ & 0.0750 & \textbf{0.0407} & \textcolor{green!50!black}{+45.7 $\uparrow\uparrow$} & 0.0633 & \textbf{0.0389} & \textcolor{green!50!black}{+38.5 $\uparrow\uparrow$} \\
$W_1$($k$NN) $\downarrow$ & 0.0039 & \textbf{0.0026} & \textcolor{green!50!black}{+33.3 $\uparrow\uparrow$} & 0.0281 & \textbf{0.0256} & \textcolor{green!50!black}{+8.9 $\uparrow$} \\
Spearman $\uparrow$ & 0.8223 & \textbf{0.8324} & \textcolor{green!50!black}{+1.2 $\uparrow$} & 0.5463 & \textbf{0.5468} & \textcolor{green!50!black}{+0.1 $\uparrow$} \\
Pearson $\uparrow$ & 0.8198 & \textbf{0.8331} & \textcolor{green!50!black}{+1.6 $\uparrow$} & 0.5452 & \textbf{0.5456} & \textcolor{green!50!black}{+0.1 $\uparrow$} \\
Shell F1 $\uparrow$ & 0.3110 & \textbf{0.4279} & \textcolor{green!50!black}{+37.6 $\uparrow\uparrow$} & \textbf{0.2134} & 0.2126 & \textcolor{red}{-0.4 $\downarrow$} \\
Edge ROC $\uparrow$ & 0.8971 & \textbf{0.9327} & \textcolor{green!50!black}{+4.0 $\uparrow$} & \textbf{0.7884} & 0.7874 & \textcolor{red}{-0.1 $\downarrow$} \\
\hline
\end{tabular}%
}
\vspace{-0.8cm}
\end{table}


\begin{figure*}[t]
  \centering
  \begin{minipage}[t]{0.48\textwidth}
    \centering
    \includegraphics[width=\linewidth]{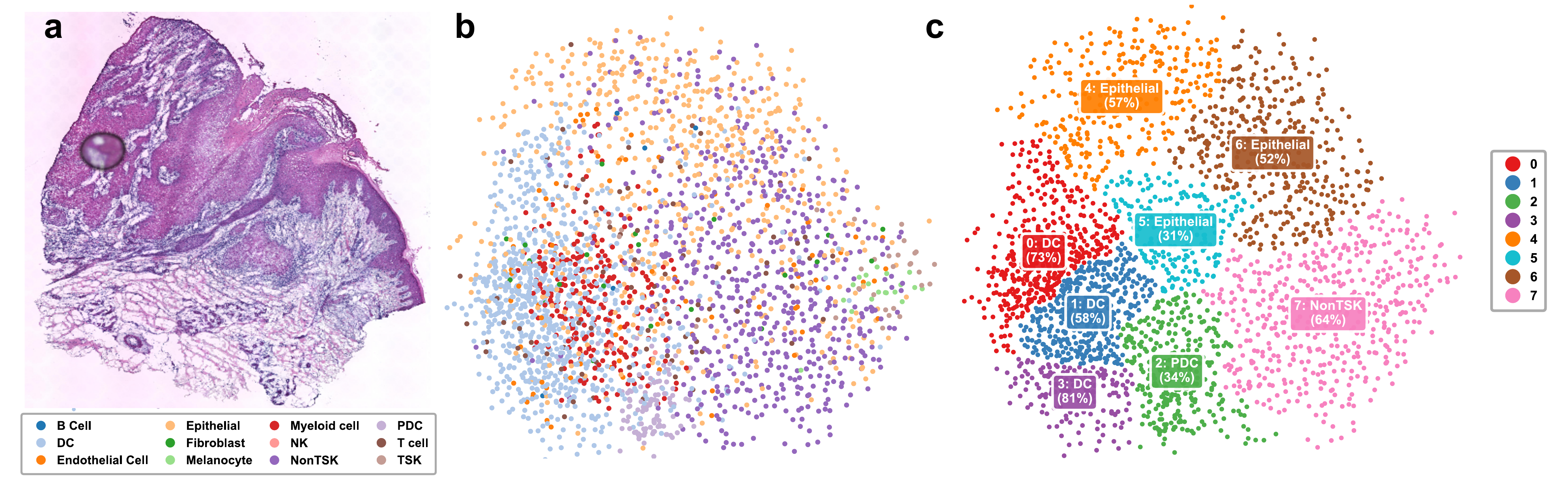}
    \vspace{-0.4cm}
    \captionsetup{justification=centering, font=small}
    \captionof{figure}{hSCC single-cell reconstruction and unsupervised spatial domains.
    (a) Reference H\&E from training ST slide.
    (b) Inferred single-cell coordinates colored by cell-type annotations.
    (c) Leiden spatial communities ($k$=15, resolution 0.2) labeled by majority cell type and purity.}
    \label{fig:hscc_domains}
    \vspace{-0.2cm}
  \end{minipage}%
  \hfill
  \begin{minipage}[t]{0.48\textwidth}
    \centering
    \includegraphics[width=\linewidth]{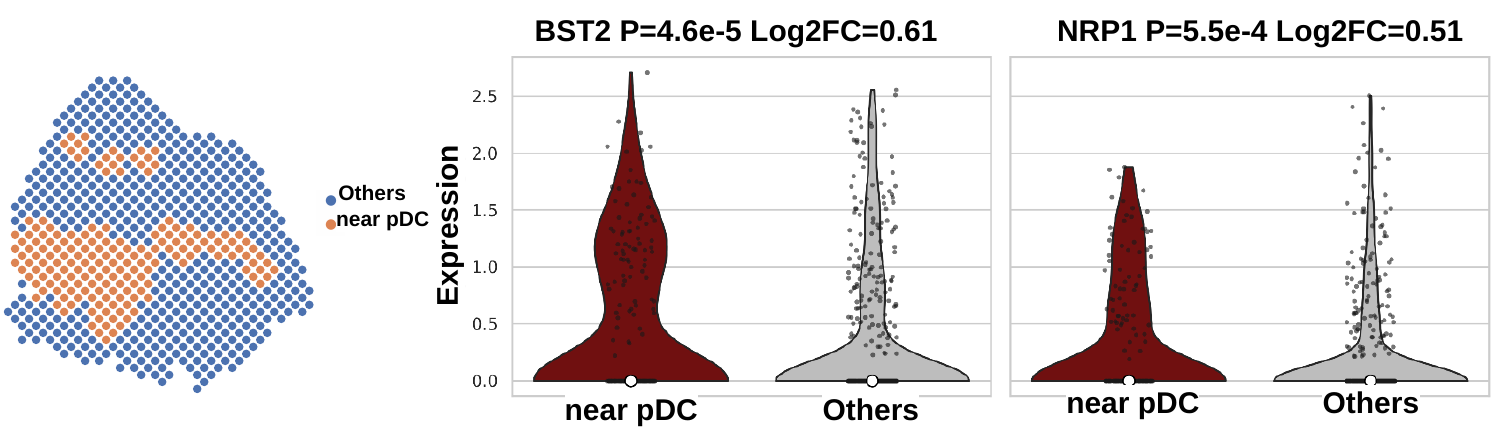}
    \vspace{-0.4cm}
    \captionsetup{justification=centering, font=small}
    \captionof{figure}{Spatial and expression validation of pDC marker genes BST2 and NRP1.
    Left: Visium spots from the hSCC reference slide; pDC-near spots (orange) defined within $1.5\times$ spot pitch of predicted pDC positions.
    Right: both markers show significantly higher expression in pDC-near regions.}
    \label{fig:pdc_markers}
    \vspace{-0.3cm}
  \end{minipage}%
\end{figure*}

\begin{figure}[H]
  \centering
  \vspace{-0.4cm}
  \includegraphics[width=\linewidth]{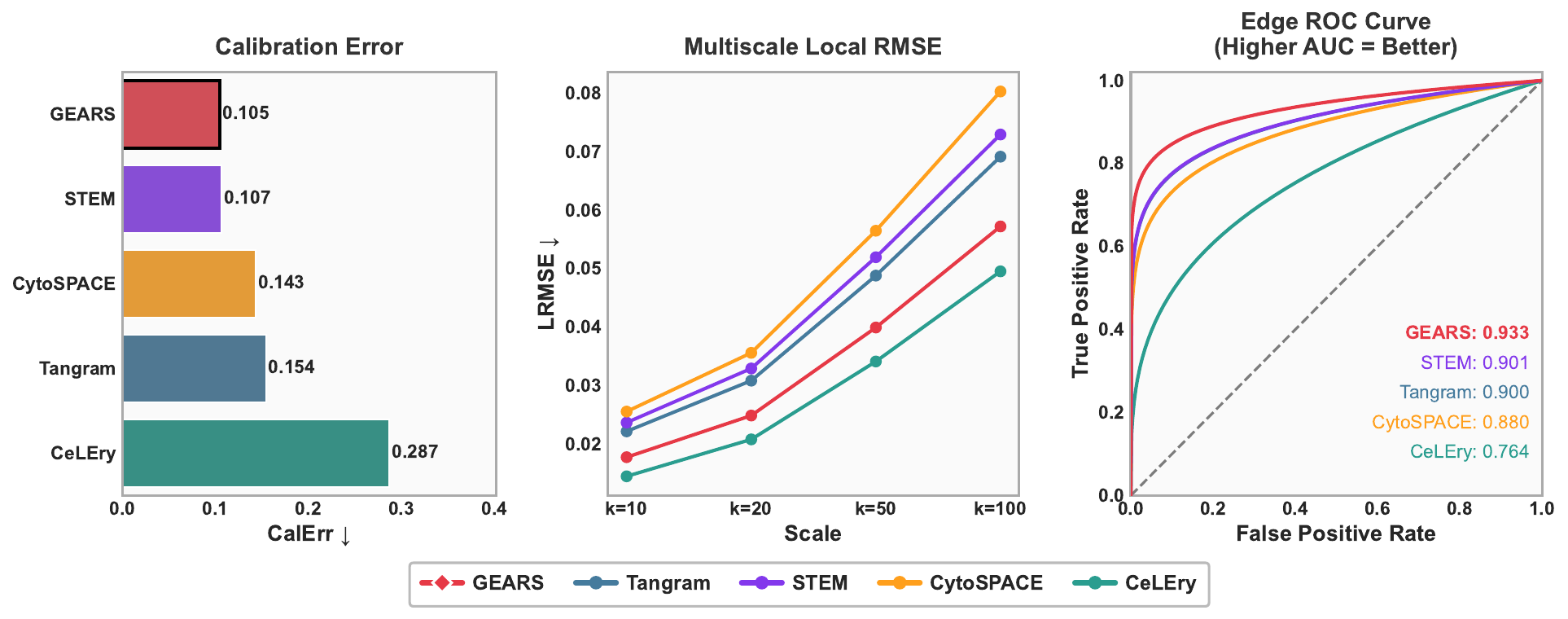}
  \vspace{-0.5cm}
  \captionsetup{justification=centering, font=small}
  \caption{Geometry fidelity diagnostics (Mouse Atlas): \textbf{Left:} calibration error, \textbf{Middle:} multiscale local RMSE at different neighborhood sizes, and \textbf{Right:} edge ROC curves (AUC). Lower is better for calibration/RMSE; higher is better for AUC.}
  \vspace{-0.4cm}
  \label{fig:geom_diagnostics}
\end{figure}

\noindent\textbf{Geometry fidelity diagnostics.}
To complement the aggregate metrics in Table~\ref{tab:benchmark}, we report three diagnostic views of reconstruction quality on the Mouse Atlas experiment (Fig.~\ref{fig:geom_diagnostics}). \emph{Calibration error} summarizes how well predicted distances match ground-truth distance magnitudes. \emph{Multiscale local RMSE} reports distance error restricted to progressively larger neighborhood radii ($k=10,20,50,100$), separating micro- vs.\ macro-local distortion. Finally, the \emph{edge ROC curve} evaluates near/far structure by treating ground-truth local pairs as positives and using predicted distances as scores. Top baselines are shown for clarity. GEARS achieves the lowest calibration error and the highest edge AUC, indicating strong distance-scale calibration and robust recovery of near/far neighborhood structure. CeLEry attains lower multiscale local RMSE, which primarily reflects absolute distance magnitude error within local radii; in contrast, GEARS emphasizes preserving neighborhood topology, consistent with its stronger edge-recovery performance.

\noindent\textbf{hSCC single-cell domain structure via unsupervised spatial communities.}
Because dissociated scRNA-seq lacks ground-truth coordinates, we evaluate hSCC reconstructions by testing whether the inferred geometry exhibits \emph{non-random spatial domain structure} consistent with coarse biology. We train on spot-based ST from the hSCC patient and infer a 2D configuration for the paired dissociated single cells from the same specimen (not slide-matched). Coloring the inferred layout by cell-type annotations reveals compartmentalization, with immune-enriched regions (e.g., DC/PDC) separated from epithelial/tumor-program regions (Epithelial/NonTSK) (Fig.~\ref{fig:hscc_domains}b). To summarize spatial structure without using labels, we build a $k$NN graph on the inferred coordinates ($k=15$) and apply Leiden community detection~\cite{Traag_2019}, yielding $K=8$ spatial communities at resolution $0.2$ (Fig.~\ref{fig:hscc_domains}c). Annotating each community by its majority cell type shows multiple DC-dominant domains (purity $58$--$81\%$), a NonTSK-enriched domain ($64\%$), and several epithelial-enriched domains ($31$--$57\%$), while mixed domains localize interfaces and heterogeneous regions. The agreement between label-free spatial communities and coarse annotations supports that the inferred single-cell geometry captures structured organization beyond trivial mixing.

\noindent\textbf{Spatially Resolved Marker Validation on hSCC.} To test whether the inferred geometry captures biologically meaningful spatial signal, we examined expression of prespecified pDC markers on the reference ST slide. Predicted pDCs in the dissociated scRNA-seq cohort were mapped to the Visium coordinate space, and spots within $1.5\times$ spot pitch of any predicted pDC position were labeled pDC-near (177 pDC-near vs 489 other spots; Fig.~\ref{fig:pdc_markers}, left). Both BST2 and NRP1 showed significantly higher expression in pDC-near regions (BST2: $P{=}4.6{\times}10^{-5}$, $\log_2\text{FC}{=}0.61$; NRP1: $P{=}5.5{\times}10^{-4}$, $\log_2\text{FC}{=}0.51$; Fig.~\ref{fig:pdc_markers}, right), confirming that the inferred single-cell geometry places pDCs in spatially coherent positions consistent with expected marker localization.

\begin{figure}[H]
  \centering
  \vspace{-0.3cm}
  \includegraphics[width=\linewidth]{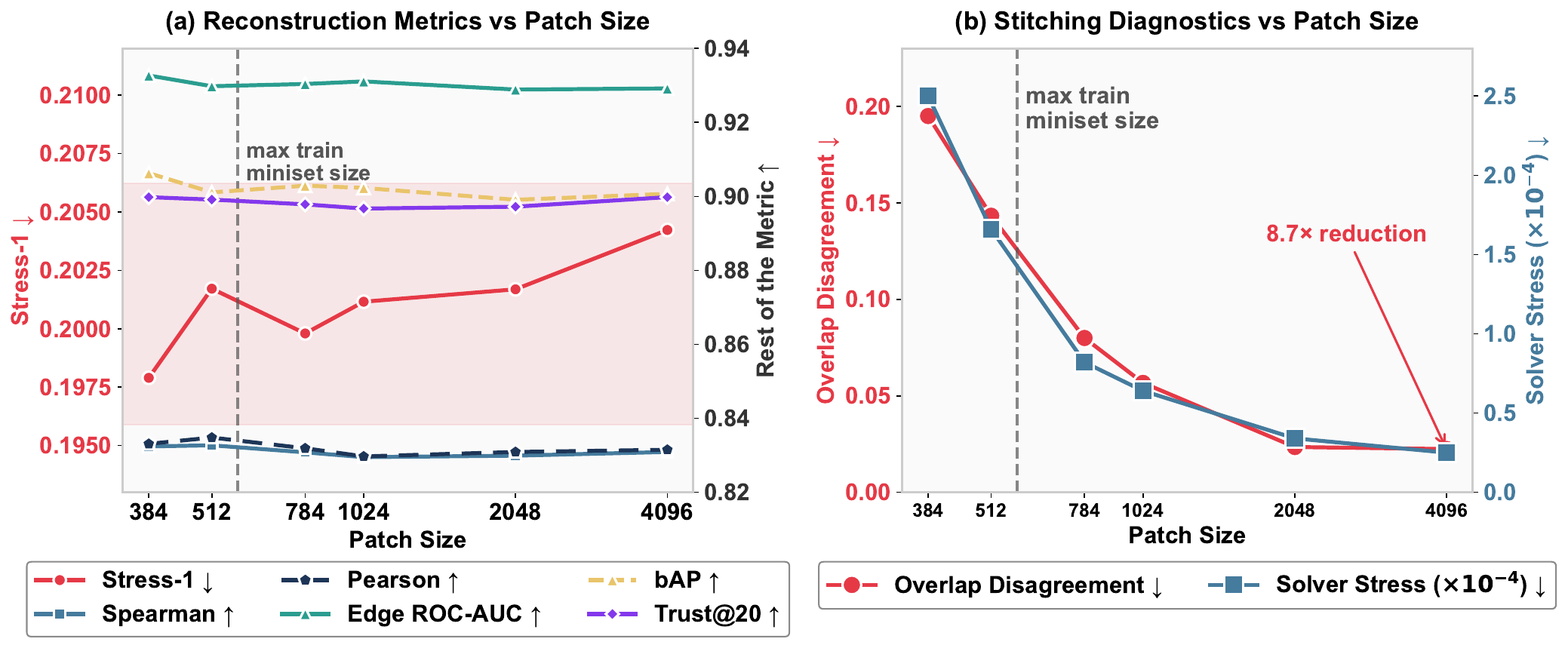}
  \vspace{-0.5cm}
  \captionsetup{justification=centering, font=small}
  \caption{Patch-size sensitivity of patchwise inference.
  Reconstruction metrics are stable across patch sizes, while stitching diagnostics improve as patches grow.}
  \vspace{-0.5cm}
  \label{fig:patch}
\end{figure}

\noindent\textbf{Patch-size sensitivity of distance-first patchwise inference.}
To test whether our patchwise pipeline is sensitive to the inference patch size, we randomly subsampled $N{=}4096$ cells from Mouse Atlas (fixed seed) and ran the inference pipeline with patch sizes $\{384,512,784,1024,2048,4096\}$, keeping all other hyperparameters fixed (Fig.~\ref{fig:patch}). Across this wide range, spanning below and above the maximum miniset size used during training—global and local reconstruction metrics remain essentially unchanged, indicating that GEARS does not rely on a specific inference cardinality and that patchwise stitching does not degrade geometric fidelity. 
In contrast, stitching becomes easier with larger patches: overlap disagreement decreases monotonically and the global solver residual drops, consistent with fewer patch boundaries and more internally consistent distance measurements per edge. 
Overall, these results support that our distance-first inference is robust to patch-size choices, with larger patches mainly improving stitching self-consistency rather than changing the reconstructed geometry.

\noindent\textbf{Sensitivity to Patch-Reliability Weighting in Distance Stitching.}
\label{app:stitching_sensitivity} We evaluate the sensitivity of distance-first stitching to the patch-reliability weights used during aggregation. We compare (i) reliability-weighted aggregation (default) against (ii) uniform aggregation ($a_p\equiv 1$), keeping patch sampling, within-patch distance extraction, and all filtering/solve settings fixed. Results are nearly unchanged, indicating that stitched geometry is not brittle to this choice and that patchwise distance predictions are already largely self-consistent. We attribute this stability to the model’s context-invariant training signals: in particular, paired overlapping minisets with overlap-consistency losses and pose-invariant Gram supervision, which encourage shared cells to receive compatible local geometries across different patch contexts; reliability weighting therefore mainly acts as a safety mechanism rather than a critical tuning knob.

\vspace{-0.1cm}
\begin{table}[!hbtp]
\centering
\small
\captionsetup{font=small}
\caption{Sensitivity of distance stitching to patch-reliability weighting ($n_{\text{patch}}=1024$, $N=4096$ cells).}
\label{tab:stitching_weights}
\vspace{-0.2cm}
\begin{tabular}{l|cc|r}
\hline
\textbf{Metric} & \textbf{Weighted} & \textbf{Uniform} & $\boldsymbol{\Delta}$\textbf{\%} \\
\hline
Stress-1 $\downarrow$ & \textbf{0.2012} & 0.2014 & +0.1 \\
Spearman $\uparrow$ & \textbf{0.8295} & 0.8291 & -0.0 \\
Edge ROC-AUC $\uparrow$ & \textbf{0.9311} & 0.9298 & -0.1 \\
bAP $\uparrow$ & \textbf{0.9023} & 0.9017 & -0.1 \\
Trust@20 $\uparrow$ & \textbf{0.8967} & 0.8962 & -0.1 \\
Cont.@20 $\uparrow$ & \textbf{0.9194} & 0.9118 & -0.8 \\
Local Stress $\downarrow$ & \textbf{0.5452} & 0.5609 & +2.9 \\
\hline
\end{tabular}
\vspace{-0.3cm}
\end{table}

\begin{figure}[H]
  \centering
  \vspace{-0.3cm}
  \includegraphics[width=\linewidth]{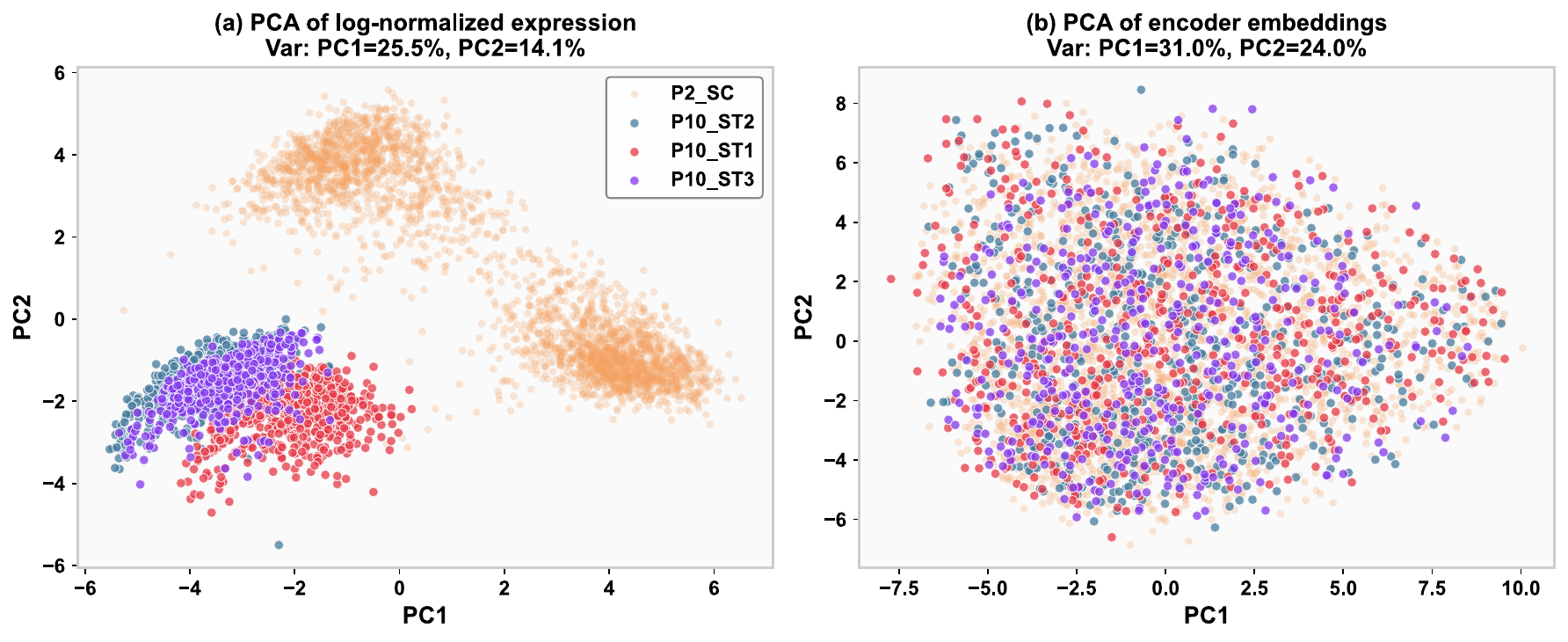}
  \vspace{-0.5cm}
  \captionsetup{justification=centering, font=small}
    \caption{Domain alignment via shared encoder. (a) Log-normalized expression separates by source. (b) Learned embeddings show cross-source mixing across ST slides (P10) and scRNA-seq (P2).}

  \vspace{-0.5cm}
\label{fig:pca_encoder}
\end{figure}

\noindent\textbf{Cross-slide and cross-cohort embedding alignment.}
We evaluate whether the shared encoder $f_\theta$ reduces domain shift across hSCC ST sections and a dissociated scRNA-seq cohort by embedding three ST slides from patient 10 (P10\_ST1--P10\_ST3) together with scRNA-seq from a different patient (P2\_SC).
Fig.~\ref{fig:pca_encoder} compares PCA on log-normalized expression versus PCA on the learned embeddings: log-normalized expression separates strongly by slide/cohort, whereas the encoder produces substantially increased cross-source mixing.
This effect is consistent with our training objective, which combines VICReg-style invariances with explicit domain-alignment losses to suppress slide/cohort-specific nuisance variation while preserving biological signal.
We quantify mixing using $k$NN domain-mixing metrics in embedding space ($k$=20), obtaining normalized neighbor entropy~\cite{haghverdi2018batch} $0.7881$ and normalized iLISI~\cite{Korsunsky461954} $0.6248$ (0=no mixing, 1=ideal mixing).

\vspace{-0.3cm}
\section{Related Work}
\textbf{Spot-level deconvolution.}
Many methods model each ST spot as a mixture of cell states and estimate spot-wise abundances using scRNA-seq as reference, without producing explicit per-cell coordinates. Probabilistic generative approaches such as Cell2location~\cite{kleshchevnikov2022cell2location}, Stereoscope~\cite{andersson2020single}, and DestVI/scvi-tools variants~\cite{lopez2022destvi, gayoso2022python} fit spot expression models and return proportions (and sometimes continuous state variation) at the spot level. Classical regression/likelihood pipelines including RCTD~\cite{cable2022robust} and SPOTlight~\cite{elosua2021spotlight} remain strong baselines. Overall, these methods explain \emph{what} is present at measured locations, but remain spot-centric and require additional post hoc rules to place individual cells in continuous 2D space.


\noindent\textbf{Cell-to-space alignment and placement.}
A second family aligns scRNA-seq profiles to ST measurements to infer cell-to-spot relations and induce spatial positions. Anchor-based transfer (e.g., Seurat)~\cite{stuart2019comprehensive, hao2021integrated} enables label/expression transfer, while Tangram~\cite{biancalani2021deep}, optimal transport methods such as SpaOTsc~\cite{cang2020inferring}, and reconstruction approaches such as novoSpaRc~\cite{moriel2021novosparc} explicitly learn couplings between cells and spatial locations. Recent learning-based methods strengthen the mapping objective with spatial-graph supervision or additional constraints (e.g., STEM~\cite{hao2024stem}, Celloc~\cite{yin2024accurate}, COME~\cite{wei2025come}), and some directly regress coordinates (e.g., scSpace~\cite{qian2023reconstruction}, CeLEry~\cite{zhang2023leveraging}). Despite differing objectives, most approaches ultimately tether inferred positions to the observed spot lattice (or convex combinations), inheriting discretization tied to the measurement grid.

\vspace{-0.25cm}
\section{Concluding Remarks}
We propose \textit{GEARS}, a geometry-first framework that reconstructs continuous single-cell spatial organization from dissociated scRNA-seq using ST as geometric supervision, without cell-type labels, histology, or explicit cell-to-spot assignment. GEARS decouples reconstruction from slide-specific coordinate frames by (i) learning a domain-invariant expression embedding that aligns ST and SC profiles, (ii) training a permutation-equivariant generator with an EDM-preconditioned residual diffusion refiner under pose-invariant Gram-based targets derived from ST coordinates, and (iii) performing distance-first, patchwise inference that stitches overlapping local reconstructions into a coherent global geometry via aggregated distance constraints and a global distance-geometry solve. Across the Mouse Atlas and multi-slide hSCC generalization benchmarks, GEARS improves global distance preservation and local neighborhood fidelity over strong mapping and deconvolution baselines while yielding coherent unsupervised spatial domains. GEARS is expected to be less reliable under very strong ST--SC domain shift, severe cell-composition mismatch between modalities, or when ST resolution provides only weak geometric supervision. Cross-patient and cross-cohort geometry generalization remains an important direction for future work.

\begin{acks}
We thank Jeovani Overstreet for helpful discussions throughout this work. This work was supported by grants from the National Science Foundation (NSF-III2246796 and NSF-III2152030).
\end{acks}

\bibliographystyle{ACM-Reference-Format}
\balance
\bibliography{refs}

\allowdisplaybreaks
\section{Appendix}
\subsection{Evaluation Metrics}
\label{app:metrics}

We evaluate on a set of $N$ points with ground-truth 2D coordinates $\mathbf{X}^{\mathrm{GT}}\in\mathbb{R}^{N\times 2}$ (rows $\mathbf{X}^{\mathrm{GT}}_{i,:}$) and ground-truth pairwise distances $\mathbf{D}^{\mathrm{GT}}\in\mathbb{R}^{N\times N}$, where $D^{\mathrm{GT}}_{ij}=\|\mathbf{X}^{\mathrm{GT}}_{i,:}-\mathbf{X}^{\mathrm{GT}}_{j,:}\|_2$. A method outputs predicted coordinates $\mathbf{X}\in\mathbb{R}^{N\times 2}$ and/or predicted distances $\mathbf{D}\in\mathbb{R}^{N\times N}$; when only $\mathbf{X}$ is available, we set $D_{ij}=\|\mathbf{X}_{i,:}-\mathbf{X}_{j,:}\|_2$.

\paragraph{Vectorization.}
Let $\operatorname{vec}_\triangle(\mathbf{D})\in\mathbb{R}^{N(N-1)/2}$ denote the vector of the upper-triangular entries $\{D_{ij}: 1\le i<j\le N\}$ (excluding the diagonal). We write $\mathbf{d}^{\mathrm{GT}}=\operatorname{vec}_\triangle(\mathbf{D}^{\mathrm{GT}})$ and $\mathbf{d}=\operatorname{vec}_\triangle(\mathbf{D})$.

\paragraph{Global geometry: Spearman and Pearson.}
Pearson correlation is
\[
\rho_P(\mathbf{D},\mathbf{D}^{\mathrm{GT}})
=
\frac{\operatorname{Cov}(\mathbf{d},\mathbf{d}^{\mathrm{GT}})}{\operatorname{Std}(\mathbf{d})\operatorname{Std}(\mathbf{d}^{\mathrm{GT}})}.
\]
Spearman correlation $\rho_S(\mathbf{D},\mathbf{D}^{\mathrm{GT}})$ is computed as Pearson correlation of the rank-transformed vectors $\operatorname{rank}(\mathbf{d})$ and $\operatorname{rank}(\mathbf{d}^{\mathrm{GT}})$.

\paragraph{Global geometry: Stress-1.}
We report Kruskal Stress-1 between pairwise distances:
\[
\operatorname{Stress\mbox{-}1}(\mathbf{D},\mathbf{D}^{\mathrm{GT}})
=
\sqrt{
\frac{\sum_{i<j}\left(D_{ij}-D^{\mathrm{GT}}_{ij}\right)^2}
{\sum_{i<j} \left(D^{\mathrm{GT}}_{ij}\right)^2}
}.
\]

\paragraph{Neighborhood scale ($R_{20}$).}
Let $\pi_i^{\mathrm{GT}}(1),\dots,\pi_i^{\mathrm{GT}}(N-1)$ be the indices of points sorted by increasing ground-truth distance from $i$ (excluding $i$). Define the per-point 20-NN radius as $r_i=D^{\mathrm{GT}}_{i,\pi_i^{\mathrm{GT}}(20)}$ and the global neighborhood radius
\[
R_{20}=\operatorname{median}_{i\in\{1,\dots,N\}}\, r_i.
\]

\paragraph{Local geometry: Edge ROC-AUC.}
We define positives and negatives using the ground-truth radius:
$\mathcal{E}^+=\{(i,j): i<j,\; D^{\mathrm{GT}}_{ij}\le R_{20}\}$ and $\mathcal{E}^-=\{(i,j): i<j,\; D^{\mathrm{GT}}_{ij}>R_{20}\}$.
We score each pair by predicted proximity $s_{ij}=-D_{ij}$ and compute ROC-AUC for discriminating $\mathcal{E}^+$ vs.\ $\mathcal{E}^-$.

\paragraph{Local geometry: balanced Average Precision (bAP).}
Using the same labels and scores as above, we compute a class-balanced AP by weighting positives and negatives equally in the precision--recall computation. One equivalent implementation is to use weighted counts with $w^+=1/|\mathcal{E}^+|$ and $w^-=1/|\mathcal{E}^-|$ when forming precision/recall along the score-sorted list of pairs.

\paragraph{Local geometry: Shell F1 (macro).}
We measure multi-scale edge preservation by defining $S_{\text{sh}}$ distance shells with radii
$0=r_0<r_1<\dots<r_{S_{\text{sh}}}$.
In our implementation, radii can be set as multiples of $R_{20}$ (e.g., $r_s = s\cdot R_{20}$) or as quantiles of $\{D^{\mathrm{GT}}_{ij}\}$; the key requirement is that shells are defined from the ground truth.
For shell $s$, define shell positives
$\mathcal{E}_s^+=\{(i,j): i<j,\; r_{s-1}<D^{\mathrm{GT}}_{ij}\le r_s\}$,
and predict shell membership using $D_{ij}$ with the same thresholds. Let $\operatorname{Prec}_s$ and $\operatorname{Rec}_s$ be precision and recall for shell $s$; the shell F1 is
$F1_s=\frac{2\,\operatorname{Prec}_s\,\operatorname{Rec}_s}{\operatorname{Prec}_s+\operatorname{Rec}_s}$,
and we report the macro average $\frac{1}{S_{\text{sh}}}\sum_{s=1}^{S_{\text{sh}}} F1_s$.

\paragraph{Local geometry: Calibration error (CalErr).}
We measure local length-scale calibration by comparing the predicted and ground-truth $k$NN radii. For a given $k$ (we use $k=20$), define the ground-truth radius
$r_i(k) = D^{\mathrm{GT}}_{i,\pi^{\mathrm{GT}}_i(k)}$ and the predicted radius using the same
GT neighbor identity $\hat{r}_i(k) = D_{i,\pi^{\mathrm{GT}}_i(k)}$. The calibration error is
\begin{equation*}
\mathrm{CalErr}(k) \;=\; \frac{1}{N}\sum_{i=1}^N \left|\frac{\hat{r}_i(k)}{r_i(k)} - 1\right|.
\end{equation*}
Lower is better (perfect calibration gives $\mathrm{CalErr}=0$).

\paragraph{Local geometry: Multiscale local RMSE (LRMSE).}
To quantify absolute distance magnitude error at different neighborhood sizes, we evaluate RMSE over GT $k$NN pairs. Let $E_k=\{(i,\pi^{\mathrm{GT}}_i(t)) : i=1..N,\; t=1..k\}$ and let
$R_k=\mathrm{median}_i\, r_i(k)$ be the typical GT scale at neighborhood size $k$.
We report the normalized multiscale local RMSE:
\begin{equation*}
\mathrm{LRMSE}(k) \;=\; \sqrt{\frac{1}{|E_k|}\sum_{(i,j)\in E_k}
\left(\frac{D_{ij}-D^{\mathrm{GT}}_{ij}}{R_k}\right)^2 }.
\end{equation*}
We plot $\mathrm{LRMSE}(k)$ for $k\in\{10,20,50,100\}$.

\paragraph{Neighborhood quality: Trustworthiness@k and Continuity@k.}
Let $r(i,j)$ be the rank of point $j$ among neighbors of $i$ sorted by increasing ground-truth distance $D^{\mathrm{GT}}_{ij}$, and let $r^{\mathrm{pred}}(i,j)$ be the analogous rank under predicted distances $D_{ij}$. Let $\mathcal{N}_k^{\mathrm{GT}}(i)=\{\pi_i^{\mathrm{GT}}(1),\dots,\pi_i^{\mathrm{GT}}(k)\}$ be the ground-truth $k$NN set and $\mathcal{N}_k^{\mathrm{pred}}(i)$ the predicted $k$NN set. Define
$U_k(i)=\mathcal{N}_k^{\mathrm{pred}}(i)\setminus \mathcal{N}_k^{\mathrm{GT}}(i)$ and
$V_k(i)=\mathcal{N}_k^{\mathrm{GT}}(i)\setminus \mathcal{N}_k^{\mathrm{pred}}(i)$.
Trustworthiness is
\[
\operatorname{Trust@}k
=
1-\frac{2}{Nk(2N-3k-1)}\sum_{i=1}^N\sum_{j\in U_k(i)}(r(i,j)-k),
\]
and continuity is
\[
\operatorname{Cont@}k
=
1-\frac{2}{Nk(2N-3k-1)}\sum_{i=1}^N\sum_{j\in V_k(i)}(r^{\mathrm{pred}}(i,j)-k).
\]
We use $k=20$.

\paragraph{Distribution: Sliced Wasserstein distance (SWD).}
After canonicalizing $\mathbf{X}^{\mathrm{GT}}$ and $\mathbf{X}$ by centering and isotropic scaling, SWD compares their spatial distributions via random 1D projections. Sample unit directions $\{\mathbf{u}_\ell\}_{\ell=1}^{L_{\text{proj}}}$ on the unit circle. For each $\ell$, project points to scalars
$p_i^{(\ell)}=\mathbf{u}_\ell^\top \mathbf{X}^{\mathrm{GT}}_{i,:}$ and
$\hat{p}_i^{(\ell)}=\mathbf{u}_\ell^\top \mathbf{X}_{i,:}$,
sort both lists, and compute the 1D Wasserstein-1 distance. SWD is the average over projections:
\[
\operatorname{SWD}(\mathbf{X},\mathbf{X}^{\mathrm{GT}})
=
\frac{1}{L_{\text{proj}}}\sum_{\ell=1}^{L_{\text{proj}}}
W_1\!\left(\{p_i^{(\ell)}\}_{i=1}^N,\{\hat{p}_i^{(\ell)}\}_{i=1}^N\right).
\]

\paragraph{Distribution: $W_1$ on $k$NN distance distributions.}
For each point $i$, collect its $k$NN distances under ground truth and prediction:
$\mathcal{S}^{\mathrm{GT}}=\{D^{\mathrm{GT}}_{i,\pi_i^{\mathrm{GT}}(t)}: i=1..N,\; t=1..k\}$ and
$\mathcal{S}=\{D_{i,\pi_i^{\mathrm{pred}}(t)}: i=1..N,\; t=1..k\}$,
where $\pi_i^{\mathrm{pred}}(t)$ is the $t$-th nearest neighbor under $\mathbf{D}$.
We report the Wasserstein-1 distance between the empirical distributions of $\mathcal{S}^{\mathrm{GT}}$ and $\mathcal{S}$, with $k=20$.


\subsection{Hyperparameters and Reproducibility}

\noindent\textbf{Shared Encoder (Stage A).}
The shared encoder is a three-layer MLP with hidden dimensions $[512, 256, 128]$ producing $h{=}128$-D embeddings. VICReg is trained with invariance and variance weights $\lambda_{\text{inv}}{=}\lambda_{\text{var}}{=}25$, covariance weight $\lambda_{\text{cov}}{=}1$, and target standard deviation $\gamma{=}1$. The two augmented views per profile are generated with stochastic gene dropout (rate $0.3$), additive Gaussian noise (std $0.015$), and multiplicative scale jitter (range $0.25$). The adversarial discriminator is a two-layer MLP with hidden width $512$ and dropout $0.1$; the GRL coefficient is ramped from $0$ to $1.0$ over $200$ epochs following a $50$-epoch warmup. Auxiliary alignment terms include a slide-adversary weight of $50$, an MMD term (weight $20$), a CORAL term (weight $1$), and a local-alignment term (weight $4$). Stage A trains for up to $2000$ epochs with AdamW (lr $10^{-4}$, batch size $32$) and is halted early by a $k$NN balanced-probe guard ($k{=}20$, margin $0.08$).

\noindent\textbf{Geometry Model (Stage C).}
The context encoder uses $3$ ISAB blocks with $n_{\text{heads}}{=}4$, $m{=}128$ inducing points, and context dimension $c{=}256$. The generator adds $4$ ISAB blocks ($m{=}128$) followed by an MLP head producing the overcomplete latent geometry of dimension $d{=}32$. The denoiser consists of $4$ blocks with per-block FiLM conditioning, $4$ attention heads, and $128$ inducing points; self-conditioning is enabled. Geometric inductive features include a $24$-bin distance bias, $8$-bin angle features, $15$-NN local graphs, and $16$ landmarks. We sample $4500$ paired minisets per epoch with overlap fraction $\alpha{=}0.5$ and minimum overlap $|I|{\geq}20$. Noise levels are sampled stratified log-uniform in $[\sigma_{\min}, \sigma_{\text{cap}}]$ under a $3$-stage curriculum (minimum $200$ epochs per stage). Loss weights are: Gram $1.0$, log-trace scale $0.5$, NCA $1.0$ ($\tau_{\text{NCA}}{=}0.5$); overlap-consistency losses are gated at $\sigma{<}1.0$. Stage C trains jointly for $300$ epochs with AdamW (lr $10^{-4}$, batch size $32$, gradient clipping).

\noindent\textbf{Patchwise Inference.}
The locality graph is constructed as a mutual-$k$NN graph with $k_Z{=}50$ and Jaccard threshold $\tau_J{=}0.2$. Patches are sampled at $10$ walks per cell with overlap fraction $0.7$ and a minimum of $25$ shared cells between neighboring patches. Each patch is denoised over $600$ EDM steps from $\sigma_{\text{start}}{=}3\sigma_{\text{data,resid}}$ to $\sigma_{\min}$. Distance stitching uses a reliability-weighted median with minimum support $M_{\min}{=}2$ patches and spread threshold $\tau_{\text{spread}}{=}0.50$. The global solve initializes coordinates with Landmark Isomap ($128$ landmarks) and runs $1000$ iterations of weighted Huber minimization ($\delta{=}0.1$, anchor weight $0.1$, lr $10^{-2}$).

\noindent\textbf{Environment.}
All experiments are conducted on Ubuntu 22.04.5 LTS with an AMD Ryzen Threadripper 2950X 16-Core Processor, 4$\times$ NVIDIA RTX A4500 GPUs (20\,GB VRAM each), and 128\,GB system RAM. The framework uses Python 3.10.12 and PyTorch 2.2.1 with CUDA 11.8 support. Training uses PyTorch Lightning Fabric with DDP via \texttt{torchrun}. All runs use fp16 mixed precision and a fixed random seed of $42$ for reproducibility.

\end{document}